\documentclass[a4paper,11pt]{article}
\usepackage{theapa, rawfonts}
\usepackage{a4wide}

\usepackage[implicit=false]{hyperref}

\usepackage{epsfig}
\usepackage[dvipsnames]{xcolor}

\usepackage{latexsym}
\usepackage{amsmath}
\usepackage{tikz}
\usepackage{tikzit}
\input{book.tikzdefs}
\tikzstyle{single}=[draw=black,text=black]

\tikzstyle{env}=[copoint,regular polygon rotate=0,minimum width=0.2cm, fill=black]

\tikzstyle{probs}=[shape=semicircle,fill=white,draw=black,shape border rotate=180,minimum width=1.2cm]

\tikzstyle{diredges}=[every to/.style={diredge}]
\tikzstyle{math matrix}=[matrix of math nodes,left delimiter=(,right delimiter=),inner sep=2pt,column sep=1em,row sep=0.5em,nodes={inner sep=0pt},text height=1.5ex, text depth=0.25ex]

\tikzstyle{gs double edge}=[double,shorten <=-1mm,shorten >=-1mm,double distance=1pt]

\tikzstyle{inline text}=[text height=1.75ex, text depth=0.3ex,yshift=0mm]
\tikzstyle{label}=[font=\small,text height=1ex, text depth=0.15ex]
\tikzstyle{left label}=[label,anchor=east,xshift=1.5mm]
\tikzstyle{right label}=[label,anchor=west,xshift=-1.5mm]

\tikzstyle{braceedge}=[decorate,decoration={brace,amplitude=2mm,raise=-1mm}]
\tikzstyle{small braceedge}=[decorate,decoration={brace,amplitude=1mm,raise=-1mm}]
\tikzstyle{parenthesis}=[decorate,decoration={parenthesis,amplitude=2mm,raise=-1mm}]

\tikzstyle{doubled}=[line width=1.6pt] 
\tikzstyle{boldedge}=[doubled,shorten <=-0.17mm,shorten >=-0.17mm]
\tikzstyle{boldedgegray}=[doubled,gray,shorten <=-0.17mm,shorten >=-0.17mm]
\tikzstyle{singleedgegray}=[gray]

\tikzstyle{semidoubled}=[line width=1.4pt] 
\tikzstyle{semiboldedgegray}=[semidoubled,gray,shorten <=-0.17mm,shorten >=-0.17mm]

\tikzstyle{boxedge}=[semiboldedgegray]

\tikzstyle{boldedgedashed}=[very thick,dashed,shorten <=-0.17mm,shorten >=-0.17mm]
\tikzstyle{vboldedgedashed}=[doubled,dashed,shorten <=-0.17mm,shorten >=-0.17mm]
\tikzstyle{left hook arrow}=[left hook-latex]
\tikzstyle{right hook arrow}=[right hook-latex]
\tikzstyle{sembracket}=[line width=0.5pt,shorten <=-0.07mm,shorten >=-0.07mm]

\tikzstyle{causal edge}=[->,thick,gray]
\tikzstyle{causal nondir}=[thick,gray]
\tikzstyle{timeline}=[thick,gray, dashed]

\tikzstyle{cedge}=[<->,thick,gray!70!white]

\tikzstyle{empty diagram}=[draw=gray!40!white,dashed,shape=rectangle,minimum width=1cm,minimum height=1cm]
\tikzstyle{empty diagram small}=[draw=gray!50!white,dashed,shape=rectangle,minimum width=0.6cm,minimum height=0.5cm]

\tikzstyle{dot}=[inner sep=0mm,minimum width=2mm,minimum height=2mm,draw,shape=circle]  
\tikzstyle{Wsquare}=[white dot, shape=regular polygon, rounded corners=0.8 mm, minimum size=3.3 mm, regular polygon sides=3, outer sep=-0.2mm]
\tikzstyle{Wsquareadj}=[white dot, shape=regular polygon, rounded corners=0.8 mm, minimum size=3.3 mm, regular polygon sides=3, outer sep=-0.2mm, regular polygon rotate=180]
\tikzstyle{ddot}=[inner sep=0mm, doubled, minimum width=2.5mm,minimum height=2.5mm,draw,shape=circle]

\tikzstyle{black dot}=[dot,fill=black]
\tikzstyle{white dot}=[dot,fill=white,text depth=-0.2mm]
\tikzstyle{white Wsquare}=[Wsquare,fill=white,text depth=-0.2mm]
\tikzstyle{white Wsquareadj}=[Wsquareadj,fill=white,text depth=-0.2mm]
\tikzstyle{green dot}=[white dot] 
\tikzstyle{gray dot}=[dot,fill=gray!40!white,text depth=-0.2mm]
\tikzstyle{red dot}=[gray dot] 

\tikzstyle{Z}=[white dot]
\tikzstyle{X}=[gray dot]

\tikzstyle{black ddot}=[ddot,fill=black]
\tikzstyle{white ddot}=[ddot,fill=white]
\tikzstyle{gray ddot}=[ddot,fill=gray!40!white]

\tikzstyle{gray edge}=[gray!60!white]

\tikzstyle{small dot}=[inner sep=0.5mm,minimum width=0pt,minimum height=0pt,draw,shape=circle]

\tikzstyle{small black dot}=[small dot,fill=black]
\tikzstyle{small white dot}=[small dot,fill=white]
\tikzstyle{small gray dot}=[small dot,fill=gray!40!white]

\tikzstyle{mbqc dot}=[small black dot]
\tikzstyle{mbqc input dot}=[small white dot]
\tikzstyle{mbqc output dot}=[small gray dot]

\tikzstyle{causal dot}=[inner sep=0.4mm,minimum width=0pt,minimum height=0pt,draw=white,shape=circle,fill=gray!40!white]

\tikzstyle{phase dimensions}=[minimum size=5mm,font=\footnotesize,rectangle,rounded corners=2mm,inner sep=0.2mm,outer sep=-2mm,scale=0.8]
\tikzstyle{dphase dimensions}=[minimum size=5mm,font=\footnotesize,rectangle,rounded corners=2.5mm,inner sep=0.2mm,outer sep=-2mm]

\tikzstyle{white phase dot}=[dot,fill=white,phase dimensions]
\tikzstyle{white phase ddot}=[ddot,fill=white,dphase dimensions]

\tikzstyle{white rect ddot}=[draw=black,fill=white,doubled,minimum size=5mm,font=\footnotesize,rectangle,rounded corners=2.5mm,inner sep=0.2mm]
\tikzstyle{gray rect ddot}=[draw=black,fill=gray!40!white,doubled,minimum size=6mm,font=\footnotesize,rectangle,rounded corners=3mm]

\tikzstyle{gray phase dot}=[dot,fill=gray!40!white,phase dimensions]
\tikzstyle{gray phase ddot}=[ddot,fill=gray!40!white,dphase dimensions]
\tikzstyle{grey phase dot}=[gray phase dot]
\tikzstyle{grey phase ddot}=[gray phase ddot]

\tikzstyle{small phase dimensions}=[minimum size=4mm,font=\tiny,rectangle,rounded corners=2mm,inner sep=0.2mm,outer sep=-2mm]
\tikzstyle{small dphase dimensions}=[minimum size=4mm,font=\tiny,rectangle,rounded corners=2mm,inner sep=0.2mm,outer sep=-2mm]

\tikzstyle{small gray phase dot}=[dot,fill=gray!40!white,small phase dimensions]
\tikzstyle{small gray phase ddot}=[ddot,fill=gray!40!white,small dphase dimensions]

\tikzstyle{small map}=[draw,shape=rectangle,minimum height=4mm,minimum width=4mm,fill=white]

\tikzstyle{cnot}=[fill=white,shape=circle,inner sep=-1.4pt]

\tikzstyle{asym hadamard}=[fill=white,draw,shape=NEbox,inner sep=0.6mm,font=\footnotesize,minimum height=4mm]
\tikzstyle{asym hadamard conj}=[fill=white,draw,shape=NWbox,inner sep=0.6mm,font=\footnotesize,minimum height=4mm]
\tikzstyle{asym hadamard dag}=[fill=white,draw,shape=SEbox,inner sep=0.6mm,font=\footnotesize,minimum height=4mm]

\tikzstyle{hadamard}=[fill=white,draw,inner sep=0.6mm,font=\footnotesize,minimum height=4mm,minimum width=4mm]
\tikzstyle{small hadamard}=[fill=white,draw,inner sep=0.6mm,minimum height=1.5mm,minimum width=1.5mm]
\tikzstyle{small hadamard rotate}=[small hadamard,rotate=45]
\tikzstyle{dhadamard}=[hadamard,doubled]
\tikzstyle{small dhadamard}=[small hadamard,doubled]
\tikzstyle{small dhadamard rotate}=[small hadamard rotate,doubled]
\tikzstyle{antipode}=[white dot,inner sep=0.3mm,font=\footnotesize]

\tikzstyle{scalar}=[diamond,draw,inner sep=0.5pt,font=\small]
\tikzstyle{dscalar}=[diamond,doubled, draw,inner sep=0.5pt,font=\small]

\tikzstyle{small box}=[rectangle,inline text,fill=white,draw,minimum height=5mm,yshift=-0.5mm,minimum width=5mm,font=\small]
\tikzstyle{small gray box}=[small box,fill=gray!30]
\tikzstyle{medium box}=[rectangle,inline text,fill=white,draw,minimum height=5mm,yshift=-0.5mm,minimum width=10mm,font=\small]
\tikzstyle{square box}=[small box] 
\tikzstyle{medium gray box}=[small box,fill=gray!30]
\tikzstyle{semilarge box}=[rectangle,inline text,fill=white,draw,minimum height=5mm,yshift=-0.5mm,minimum width=12.5mm,font=\small]
\tikzstyle{large box}=[rectangle,inline text,fill=white,draw,minimum height=5mm,yshift=-0.5mm,minimum width=15mm,font=\small]
\tikzstyle{large gray box}=[small box,fill=gray!30]

\tikzstyle{Bayes box}=[rectangle,fill=black,draw, minimum height=3mm, minimum width=3mm]

\tikzstyle{gray square point}=[small box,fill=gray!50]

\tikzstyle{dphase box white}=[dhadamard]
\tikzstyle{dphase box gray}=[dhadamard,fill=gray!50!white]
\tikzstyle{phase box white}=[hadamard]
\tikzstyle{phase box gray}=[hadamard,fill=gray!50!white]

\tikzstyle{point}=[regular polygon,regular polygon sides=3,draw,scale=0.75,inner sep=-0.5pt,minimum width=9mm,fill=white,regular polygon rotate=180]
\tikzstyle{copoint}=[regular polygon,regular polygon sides=3,draw,scale=0.75,inner sep=-0.5pt,minimum width=9mm,fill=white]
\tikzstyle{dpoint}=[point,doubled]
\tikzstyle{dcopoint}=[copoint,doubled]

\tikzstyle{wide copoint}=[fill=white,draw,shape=isosceles triangle,shape border rotate=90,isosceles triangle stretches=true,inner sep=0pt,minimum width=1.5cm,minimum height=6.12mm]
\tikzstyle{wide point}=[fill=white,draw,shape=isosceles triangle,shape border rotate=-90,isosceles triangle stretches=true,inner sep=0pt,minimum width=1.5cm,minimum height=6.12mm,yshift=-0.0mm]
\tikzstyle{wide point plus}=[fill=white,draw,shape=isosceles triangle,shape border rotate=-90,isosceles triangle stretches=true,inner sep=0pt,minimum width=1.74cm,minimum height=7mm,yshift=-0.0mm]

\tikzstyle{wide dpoint}=[fill=white,doubled,draw,shape=isosceles triangle,shape border rotate=-90,isosceles triangle stretches=true,inner sep=0pt,minimum width=1.5cm,minimum height=6.12mm,yshift=-0.0mm]

\tikzstyle{tinypoint}=[regular polygon,regular polygon sides=3,draw,scale=0.55,inner sep=-0.15pt,minimum width=6mm,fill=white,regular polygon rotate=180] 

\tikzstyle{white point}=[point]
\tikzstyle{white dpoint}=[dpoint]
\tikzstyle{green point}=[white point] 
\tikzstyle{white copoint}=[copoint]
\tikzstyle{gray point}=[point,fill=gray!40!white]
\tikzstyle{gray dpoint}=[gray point,doubled]
\tikzstyle{red point}=[gray point] 
\tikzstyle{gray copoint}=[copoint,fill=gray!40!white]
\tikzstyle{gray dcopoint}=[gray copoint,doubled]

\tikzstyle{white point guide}=[regular polygon,regular polygon sides=3,font=\scriptsize,draw,scale=0.65,inner sep=-0.5pt,minimum width=9mm,fill=white,regular polygon rotate=180]

\tikzstyle{tiny gray point}=[tinypoint,fill=gray!40!white]

\tikzstyle{diredge}=[->]
\tikzstyle{ddiredge}=[<->]
\tikzstyle{rdiredge}=[<-]
\tikzstyle{thickdiredge}=[->, very thick]
\tikzstyle{pointer edge}=[->,very thick,gray]
\tikzstyle{pointer edge part}=[very thick,gray]
\tikzstyle{dashed edge}=[dashed]
\tikzstyle{thick dashed edge}=[very thick,dashed]
\tikzstyle{thick gray dashed edge}=[thick dashed edge,gray!40]
\tikzstyle{gray dashed edge}=[dashed edge,gray!50]
\tikzstyle{thick map edge}=[very thick,|->]

\tikzstyle{cloud}=[shape=cloud,draw,minimum width=1.5cm,minimum height=1.5cm]

\tikzstyle{map}=[draw,shape=NEbox,inner sep=2pt,minimum height=6mm,fill=white]
\tikzstyle{dashedmap}=[draw,dashed,shape=NEbox,inner sep=2pt,minimum height=6mm,fill=white]
\tikzstyle{mapdag}=[draw,shape=SEbox,inner sep=2pt,minimum height=6mm,fill=white]
\tikzstyle{mapadj}=[draw,shape=SEbox,inner sep=2pt,minimum height=6mm,fill=white]
\tikzstyle{maptrans}=[draw,shape=SWbox,inner sep=2pt,minimum height=6mm,fill=white]
\tikzstyle{mapconj}=[draw,shape=NWbox,inner sep=2pt,minimum height=6mm,fill=white]

\tikzstyle{medium map}=[draw,shape=NEbox,inner sep=2pt,minimum height=6mm,fill=white,minimum width=7mm]
\tikzstyle{medium map dag}=[draw,shape=SEbox,inner sep=2pt,minimum height=6mm,fill=white,minimum width=7mm]
\tikzstyle{medium map adj}=[draw,shape=SEbox,inner sep=2pt,minimum height=6mm,fill=white,minimum width=7mm]
\tikzstyle{medium map trans}=[draw,shape=SWbox,inner sep=2pt,minimum height=6mm,fill=white,minimum width=7mm]
\tikzstyle{medium map conj}=[draw,shape=NWbox,inner sep=2pt,minimum height=6mm,fill=white,minimum width=7mm]
\tikzstyle{semilarge map}=[draw,shape=NEbox,inner sep=2pt,minimum height=6mm,fill=white,minimum width=9.5mm]
\tikzstyle{semilarge map trans}=[draw,shape=SWbox,inner sep=2pt,minimum height=6mm,fill=white,minimum width=9.5mm]
\tikzstyle{semilarge map adj}=[draw,shape=SEbox,inner sep=2pt,minimum height=6mm,fill=white,minimum width=9.5mm]
\tikzstyle{semilarge map dag}=[draw,shape=SEbox,inner sep=2pt,minimum height=6mm,fill=white,minimum width=9.5mm]
\tikzstyle{semilarge map conj}=[draw,shape=NWbox,inner sep=2pt,minimum height=6mm,fill=white,minimum width=9.5mm]
\tikzstyle{large map}=[draw,shape=NEbox,inner sep=2pt,minimum height=6mm,fill=white,minimum width=12mm]
\tikzstyle{large map dag}=[draw,shape=SEbox,inner sep=2pt,minimum height=6mm,fill=white,minimum width=12mm]
\tikzstyle{large map conj}=[draw,shape=NWbox,inner sep=2pt,minimum height=6mm,fill=white,minimum width=12mm]
\tikzstyle{very large map}=[draw,shape=NEbox,inner sep=2pt,minimum height=6mm,fill=white,minimum width=17mm]

\tikzstyle{medium dmap}=[draw,doubled,shape=NEbox,inner sep=2pt,minimum height=6mm,fill=white,minimum width=7mm]
\tikzstyle{medium dmap dag}=[draw,doubled,shape=SEbox,inner sep=2pt,minimum height=6mm,fill=white,minimum width=7mm]
\tikzstyle{medium dmap adj}=[draw,doubled,shape=SEbox,inner sep=2pt,minimum height=6mm,fill=white,minimum width=7mm]
\tikzstyle{medium dmap trans}=[draw,doubled,shape=SWbox,inner sep=2pt,minimum height=6mm,fill=white,minimum width=7mm]
\tikzstyle{medium dmap conj}=[draw,doubled,shape=NWbox,inner sep=2pt,minimum height=6mm,fill=white,minimum width=7mm]
\tikzstyle{semilarge dmap}=[draw,doubled,shape=NEbox,inner sep=2pt,minimum height=6mm,fill=white,minimum width=9.5mm]
\tikzstyle{semilarge dmap trans}=[draw,doubled,shape=SWbox,inner sep=2pt,minimum height=6mm,fill=white,minimum width=9.5mm]
\tikzstyle{semilarge dmap adj}=[draw,doubled,shape=SEbox,inner sep=2pt,minimum height=6mm,fill=white,minimum width=9.5mm]
\tikzstyle{semilarge dmap dag}=[draw,doubled,shape=SEbox,inner sep=2pt,minimum height=6mm,fill=white,minimum width=9.5mm]
\tikzstyle{semilarge dmap conj}=[draw,doubled,shape=NWbox,inner sep=2pt,minimum height=6mm,fill=white,minimum width=9.5mm]
\tikzstyle{large dmap}=[draw,doubled,shape=NEbox,inner sep=2pt,minimum height=6mm,fill=white,minimum width=12mm]
\tikzstyle{large dmap conj}=[draw,doubled,shape=NWbox,inner sep=2pt,minimum height=6mm,fill=white,minimum width=12mm]
\tikzstyle{large dmap trans}=[draw,doubled,shape=SWbox,inner sep=2pt,minimum height=6mm,fill=white,minimum width=12mm]
\tikzstyle{large dmap adj}=[draw,doubled,shape=SEbox,inner sep=2pt,minimum height=6mm,fill=white,minimum width=12mm]
\tikzstyle{large dmap dag}=[draw,doubled,shape=SEbox,inner sep=2pt,minimum height=6mm,fill=white,minimum width=12mm]
\tikzstyle{very large dmap}=[draw,doubled,shape=NEbox,inner sep=2pt,minimum height=6mm,fill=white,minimum width=19.5mm]
\tikzstyle{huge dmap}=[draw,doubled,shape=SEbox,inner sep=2pt,minimum height=6mm,fill=white,minimum width=30mm]

\tikzstyle{muxbox}=[draw,shape=rectangle,minimum height=3mm,minimum width=3mm,fill=white]
\tikzstyle{dmuxbox}=[muxbox,doubled]

\tikzstyle{box}=[draw,shape=rectangle,inner sep=2pt,minimum height=6mm,minimum width=6mm,fill=white]
\tikzstyle{dbox}=[draw,doubled,shape=rectangle,inner sep=2pt,minimum height=6mm,minimum width=6mm,inline text,fill=white]
\tikzstyle{dmap}=[draw,doubled,shape=NEbox,inner sep=2pt,minimum height=6mm,fill=white]
\tikzstyle{dmapdag}=[draw,doubled,shape=SEbox,inner sep=2pt,minimum height=6mm,fill=white]
\tikzstyle{dmapadj}=[draw,doubled,shape=SEbox,inner sep=2pt,minimum height=6mm,fill=white]
\tikzstyle{dmaptrans}=[draw,doubled,shape=SWbox,inner sep=2pt,minimum height=6mm,fill=white]
\tikzstyle{dmapconj}=[draw,doubled,shape=NWbox,inner sep=2pt,minimum height=6mm,fill=white]

\tikzstyle{ddmap}=[draw,doubled,dashed,shape=NEbox,inner sep=2pt,minimum height=6mm,fill=white]
\tikzstyle{ddmapdag}=[draw,doubled,dashed,shape=SEbox,inner sep=2pt,minimum height=6mm,fill=white]
\tikzstyle{ddmapadj}=[draw,doubled,dashed,shape=SEbox,inner sep=2pt,minimum height=6mm,fill=white]
\tikzstyle{ddmaptrans}=[draw,doubled,dashed,shape=SWbox,inner sep=2pt,minimum height=6mm,fill=white]
\tikzstyle{ddmapconj}=[draw,doubled,dashed,shape=NWbox,inner sep=2pt,minimum height=6mm,fill=white]

\tikzstyle{smap}=[draw,shape=sNEbox,fill=white]
\tikzstyle{smapdag}=[draw,shape=sSEbox,fill=white]
\tikzstyle{smapadj}=[draw,shape=sSEbox,fill=white]
\tikzstyle{smaptrans}=[draw,shape=sSWbox,fill=white]
\tikzstyle{smapconj}=[draw,shape=sNWbox,fill=white]

\tikzstyle{dsmap}=[draw,dashed,shape=sNEbox,fill=white]
\tikzstyle{dsmapdag}=[draw,dashed,shape=sSEbox,fill=white]
\tikzstyle{dsmaptrans}=[draw,dashed,shape=sSWbox,fill=white]
\tikzstyle{dsmapconj}=[draw,dashed,shape=sNWbox,fill=white]

\tikzstyle{mmap}=[draw,shape=mNEbox]
\tikzstyle{mmapdag}=[draw,shape=mSEbox]
\tikzstyle{mmaptrans}=[draw,shape=mSWbox]
\tikzstyle{mmapconj}=[draw,shape=mNWbox]

\tikzstyle{mmapgray}=[draw,fill=gray!40!white,shape=mNEbox]
\tikzstyle{smapgray}=[draw,fill=gray!40!white,shape=sNEbox]

\tikzstyle{kpoint common}=[draw,fill=white,inner sep=1pt,minimum height=4mm]
\tikzstyle{kpoint sc}=[shape=cornerpoint,kpoint common]
\tikzstyle{kpoint adjoint sc}=[shape=cornercopoint,kpoint common]
\tikzstyle{kpoint}=[shape=cornerpoint,shorten left=5pt,kpoint common]
\tikzstyle{kpoint adjoint}=[shape=cornercopoint,shorten left=5pt,kpoint common]
\tikzstyle{kpoint conjugate}=[shape=cornerpoint,shorten right=5pt,kpoint common]
\tikzstyle{kpoint transpose}=[shape=cornercopoint,shorten right=5pt,kpoint common]
\tikzstyle{kpoint symm}=[shape=cornerpoint,shorten left=5pt,shorten right=5pt,kpoint common]

\tikzstyle{kpointdag}=[kpoint adjoint]
\tikzstyle{kpointadj}=[kpoint adjoint]
\tikzstyle{kpointconj}=[kpoint conjugate]
\tikzstyle{kpointtrans}=[kpoint transpose]

\tikzstyle{big kpoint}=[kpoint, minimum width=1.2 cm, minimum height=8mm, inner sep=4pt, text depth=3mm]

\tikzstyle{wide kpoint}=[kpoint, minimum width=1 cm, inner sep=2pt]
\tikzstyle{wide kpointdag}=[kpointdag, minimum width=1 cm, inner sep=2pt]
\tikzstyle{wide kpointconj}=[kpointconj, minimum width=1 cm, inner sep=2pt]
\tikzstyle{wide kpointtrans}=[kpointtrans, minimum width=1 cm, inner sep=2pt]

\tikzstyle{gray kpoint}=[kpoint,fill=gray!50!white]
\tikzstyle{gray kpointdag}=[kpointdag,fill=gray!50!white]
\tikzstyle{gray kpointadj}=[kpointadj,fill=gray!50!white]
\tikzstyle{gray kpointconj}=[kpointconj,fill=gray!50!white]
\tikzstyle{gray kpointtrans}=[kpointtrans,fill=gray!50!white]

\tikzstyle{gray dkpoint}=[kpoint,fill=gray!50!white,doubled]
\tikzstyle{gray dkpointdag}=[kpointdag,fill=gray!50!white,doubled]
\tikzstyle{gray dkpointadj}=[kpointadj,fill=gray!50!white,doubled]
\tikzstyle{gray dkpointconj}=[kpointconj,fill=gray!50!white,doubled]
\tikzstyle{gray dkpointtrans}=[kpointtrans,fill=gray!50!white,doubled]

\tikzstyle{white label}=[draw,fill=white,rectangle,inner sep=0.7 mm]
\tikzstyle{gray label}=[draw,fill=gray!50!white,rectangle,inner sep=0.7 mm]
\tikzstyle{black label}=[draw,fill=black,rectangle,inner sep=0.7 mm]

\tikzstyle{dkpoint}=[kpoint,doubled]
\tikzstyle{wide dkpoint}=[wide kpoint,doubled]
\tikzstyle{dkpointdag}=[kpoint adjoint,doubled]
\tikzstyle{wide dkpointdag}=[wide kpointdag,doubled]
\tikzstyle{dkcopoint}=[kpoint adjoint,doubled]
\tikzstyle{dkpointadj}=[kpoint adjoint,doubled]
\tikzstyle{dkpointconj}=[kpoint conjugate,doubled]
\tikzstyle{dkpointtrans}=[kpoint transpose,doubled]

\tikzstyle{kscalar}=[kpoint common, shape=EBox, inner xsep=-1pt, inner ysep=3pt,font=\small]
\tikzstyle{kscalarconj}=[kpoint common, shape=WBox, inner xsep=-1pt, inner ysep=3pt,font=\small]

\tikzstyle{spekpoint}=[kpoint sc,minimum height=5mm,inner sep=3pt]
\tikzstyle{spekcopoint}=[kpoint adjoint sc,minimum height=5mm,inner sep=3pt]

\tikzstyle{dspekpoint}=[spekpoint,doubled]
\tikzstyle{dspekcopoint}=[spekcopoint,doubled]

\tikzstyle{vertex}=[inner sep=0.5mm, minimum size=0pt, shape=circle, draw=black, fill=black]
\tikzstyle{vertex set}=[inner sep=0.5mm, minimum size=0pt, shape=circle, draw=black, fill=white, font={\footnotesize\boldmath}]

 \tikzstyle{upground}=[circuit ee IEC,thick,ground,rotate=90,scale=2.7]
 \tikzstyle{downground}=[circuit ee IEC,thick,ground,rotate=-90,scale=2.5]
 \tikzstyle{rightground}=[circuit ee IEC,thick,ground,rotate=0,scale=2.5]
\tikzstyle{bigground}=[regular polygon,regular polygon sides=3,draw=gray,scale=0.50,inner sep=-0.5pt,minimum width=10mm,fill=gray]

\tikzstyle{arrs}=[-latex,font=\small,auto]
\tikzstyle{arrow plain}=[arrs]
\tikzstyle{arrow dashed}=[dashed,arrs]
\tikzstyle{arrow bold}=[very thick,arrs]
\tikzstyle{arrow hide}=[draw=white!0,-]
\tikzstyle{arrow reverse}=[latex-]

\tikzstyle{none}=[fill=none]
\tikzstyle{title}=[fill=none]
\tikzstyle{lab}=[fill=white, font=\tiny]
\tikzstyle{new style 0}=[fill=white, draw=black, shape=trapezium]
\tikzstyle{new style 1}=[fill=white, draw=black, shape=circle]
\tikzstyle{new style 2}=[fill=white, draw=green, shape=circle]
\tikzstyle{new style 3}=[fill=white, draw=red, shape=circle]
\tikzstyle{new style 4}=[fill=white, draw=black, thick, shape=regular polygon, regular polygon sides=3]
\tikzstyle{element}=[fill=white, draw=black, shape=trapezium]
\tikzstyle{functional}=[fill=white, draw=black, shape=trapezium,shape border rotate=180]
\tikzstyle{counit}=[fill=black, shape=circle]
\tikzstyle{bigClasp}=[fill=white, draw=black, shape=circle, minimum width=1.7cm]

\tikzstyle{downArrow}=[none]
\tikzstyle{thickArr}=[line width = 2pt]
\tikzstyle{thick}=[line width = 2pt]
\tikzstyle{thicker}=[line width = 4pt]
\tikzstyle{every loop}=[]

\usepackage{csquotes}

\usepackage{float}

\usepackage{natbib}
    
\usepackage{enumitem}
\usepackage{braket}
\usepackage{subcaption}
\usepackage{amssymb}

\DeclareMathOperator{\Tr}{Tr}

\usepackage{graphicx}
\usepackage{amsmath}
\usepackage{amssymb}
\usepackage[detect-all]{siunitx}   
\usepackage{tabularx}
\usepackage{multirow}
\usepackage{booktabs}
\usepackage{diagbox}
\usepackage{mathtools}

\title{Quantum Methods for Managing\\Ambiguity in Natural Language Processing}

\def\email#1{{\tt\small #1}}

\author{Jurek Eisinger \email{ jurek.e@icloud.com} \\
        Ward Gauderis \email{ ward.gauderis@vub.be} \\
        Lin de Huybrecht \email{ lin.de.huybrecht@vub.be} \\
        Geraint A.\ Wiggins \email{ geraint.wiggins@vub.be} \\
       \normalsize Computational Creativity Lab,\\
       \normalsize Artificial Intelligence Research Group\\
       \normalsize Vrije Universiteit Brussel, Pleinlaan 9, 
       1050 Elsene, Belgium}

\date{}

\usepackage{todonotes} 

\begin{document}
\renewcommand{\floatpagefraction}{.95}

\maketitle
\begin{abstract}
\noindent
The 
Categorical Compositional Distributional (\textit{DisCoCat}) framework models meaning in natural language using the mathematical framework of quantum theory, expressed as formal diagrams. DisCoCat diagrams 
can be associated with tensor networks and quantum circuits. 
DisCoCat diagrams have been connected to density matrices in various contexts in Quantum Natural Language Processing (QNLP). 
Previous use of density matrices in QNLP entails modelling ambiguous words as probability distributions over more basic words (the word \texttt{queen}, e.g., might mean the reigning queen or the chess piece). 
In this article, we investigate using probability distributions over processes to account for syntactic ambiguity in sentences. The meanings of these sentences are represented by density matrices. 
We show how to create  probability distributions on quantum circuits that represent the meanings of sentences and  explain 
how this approach generalises tasks from the literature. We conduct an experiment to validate the proposed theory. 
\end{abstract}


\section{Introduction}
\label{sec:introductionPaper}


The fast-growing field of Quantum Natural Language Processing  \citep[QNLP:][]{qnlpPaper}, in which the current article is situated, seeks to explain how information is processed in human language, using the mathematical framework of quantum theory. 
In QNLP, machine learning models are quantum circuits, which capture the meaning of sentences or other pieces of linguistic information.
These models reflect an inherently compositional approach, in contrast to  state-of-the-art  machine learning models, such as deep neural networks, which renders them more interpretable \citep{coecke2020foundations}. 



The contributions of the current article are formulated in terms of diagrams in the \emph{Categorical Compositional Distributional} (\emph{DisCoCat}) framework \citep{discocat}, in which word meanings are represented by tensors of various ranks. A noun, for example, is represented by a vector, whereas an intransitive verb is modelled as a matrix, and a transitive verb is represented by a rank-three tensor. The interaction of the meaning of these words, which results in the meaning of a sentence, is guided by the \emph{pregroup grammar}. This combination of grammar and mathematical methods from tensor calculus allows the DisCoCat model to account for both the \emph{distributional} and the \emph{compositional} aspect of language. 
This connection is formally established via the mathematical framework of \emph{Category Theory}.

Originally, the DisCoCat model arose from \emph{Categorical Quantum Mechanics} \citep{categoricalQM}. This underlying connection allows the application of quantum-theoretical concepts to the modelling of language. 
One such concept is the {\it density matrix}, which has been applied in numerous contexts in QNLP \citep{densMat, word2dm, coecke2020foundations, meaningUpdatingDensMat}. A density matrix is a probability distribution over quantum states.
Further, this underlying association with Quantum Theory allows the representation of diagrams in the DisCoCat framework, which models sentence meaning in terms of tensor networks and quantum circuits, ultimately resulting in the \textit{Quantum Machine Learning} (QML) models mentioned above. 


In the current article, we extend the previous use of quantum probability distributions in QNLP to account for ambiguity with respect to different syntactic constructs. 
Consider the following pair of example sentences \citep[after][]{pronounResolution}: 
\begin{equation}
    \label{sentence:DogBrokeVase}
    \texttt{The dog broke the vase. It was clumsy.}
\end{equation}
While there are three possible readings (either the \texttt{dog}, the \texttt{vase}, or the event of the \texttt{dog} breaking the \texttt{vase} can be \texttt{clumsy}), in the following only the first two readings are considered. 

The human mind seems to maintain multiple interpretations in parallel, and may favour some over others. 
Our goal is to design a model which assigns probabilities to these individual readings. 
The resulting probability distributions can then be updated (modelling the disambiguation of linguistic meaning upon obtaining context) using updating mechanisms of density matrices \citep{meaningUpdatingDensMat}. 

From a technical perspective, we associate the two realisations of Example \ref{sentence:DogBrokeVase} with two \textit{controlled operations} on the quantum circuit representing the meanings of the words in the sentence. These controlled operations can be associated with the functions of individual words in sentences, that cause ambiguity in the sentence (in Example \ref{sentence:DogBrokeVase}, the word \texttt{it}). Thus, the function of certain words in sentences are reinterpreted from being states to being processes that alter the meaning of the ultimate sentence, by connecting words in sentences in alternative ways. 

An important point is that the human mind can reason about the likelihood of each of the realisations being correct, based only on the meanings of the words \texttt{dog}, \texttt{vase}, and \texttt{clumsy}. 
We argue that the result of explicitly training quantum circuits in several former approaches \citep[e.g.,][]{pronounResolution, qnlpInPractice} can be achieved in a more general manner when training a model on semantics, with a subsequent reasoning process (addressed in Section \ref{sec:ReasoningProcessPaper}).

In Section \ref{sec:relatedWorkPaper}, we mention related work that the current research builds on. 
In Subsection \ref{sec:linguisticsAndQCPaper}, we give an overview of the background knowledge, such as the DisCoCat framework, quantum computing and their relation. We address density matrices and their application in QNLP and compare alternative mathematical tools to model copying mechanisms of linguistic information. 
In Section \ref{sec:ambiguitySyntaxPaper}, we give several examples on how the above discussed probability distributions are created. We furthermore formulate our contribution in the \emph{DisCoCirc} \citep{DisCoCirc} framework, in which sentences are \emph{processes} altering the meaning of the words that make it up. 

In Section \ref{sec:implementation}, an experiment is described, using a dataset inspired by \citet{pronounResolution}. 
We experiment on simulations of \textit{Noisy Intermediate Scale Quantum era} (NISQ) hardware, using the \texttt{Lambeq} \citep{lambeq} library for \texttt{python}. In particular, the model that we use to train will be implemented using the \texttt{Tket} compiler, a tool to optimise and manipulate quantum circuits\footnote{\url{https://tket.quantinuum.com/api-docs/getting_started.html} (accessed on 14.10.2024)}.
Concretely, this paper makes four contributions: 
\begin{enumerate}
    \item We relate the process of copying information by modelling projections from Fock spaces, introduced by \citet{pronounResolution}, to pregroup string diagrams with Frobenius algebras, introduced below.
    \item We propose a way of modelling probability distributions over different syntactic sentences in the DisCoCat framework, together with the theory of how this is mapped to quantum circuits.
    \item We relate the proposed theory to the \texttt{DisCoCirc} framework \citep{DisCoCirc}, in which we subsequently move from pure to mixed processes representing sentences.
    \item We implement a proof of concept, where probability distributions over sentence meanings are created using quantum circuits. 
\end{enumerate}

\section{Related Work}
\label{sec:relatedWorkPaper}
\subsection{Quantum Natural Language Processing}
Much current research in natural language processing (NLP) focuses on connectionist methods, particularly those using large language models (LLMs). These models have achieved significant advances by processing massive datasets, and generalising from the structure in the data. However, they are black box models and contribute little to understanding of how language works in a human.
In contrast, applying quantum theory to NLP, giving rise to the emerging field of Quantum Natural Language Processing (QNLP), is a novel and largely unexplored approach. It, unlike LLMs, is a white box model, specifically because of the linear compositionality of the quantum operators used for the modelling. Thus, it offers the possibility of directly interpretable models, which may then be more easily comparable with human cognition. Therefore, we restrict our survey to QNLP, because comparison with LLM technology at this early stage in the development of QNLP is not useful.


\subsection{Theoretical Foundations of QNLP}

\citet{discocat} introduce the Categorical Distributional Compositional (DisCoCat) model of language. 
Based on this model, \citet{qnlpInPractice} define and explain how the training procedure of a quantum circuit representing word meanings works. In the implementation part of this work, we use the same training pipeline. 

\citet{DisCoCirc} introduces a novel linguistic modelling framework, in which words and sentences are represented as density matrices. We explain this indetail in Section \ref{sec:densityMatricesPaper}. 
Subsequently, the use of density matrices in QNLP is well explored. 
\citet{coecke2020foundations} use density matrices to model ambiguity. 
\citet{word2dm} introduce a procedure to learn density matrix embeddings, rather than word embeddings, such as the \texttt{word2vec} framework. 
\citet{chalmers} explicitly constructs density matrices on quantum circuits. 
\citet{Bruhn_2022} shows how to automatically learn explicit density matrix representations on a quantum circuit.  

Category theory provides a mathematical framework for the expression of grammatical (syntactic) stuctures. The category of \textit{completely positive maps} \citep[\texttt{CPM}:][]{densMat} formally allows density matrices to represent word meanings, because it is {\it compact closed} \citep{densMat}, which allows grammar to guide the composition of word meanings.

The diagrammatic notation for quantum systems, presented by \citet{coeckebook}, and is useful in making the current discussion more readable.
Diagrammatically, \citet{doublingPaper} introduce density matrices as \emph{thick} wires in string diagrams. 
\citet{discardingPaper} introduce a \emph{discard} map that allows a quantum system to be \enquote{ignored}. These methods will be used in the current article. 

\citet{Wijnholds_2020, pronounResolution} present  methods of modelling verb phrase ellipsis, using extensions to the Lambek calculus \citep{orgLambek}. 
For this, while \citet{Wijnholds_2020} introduces \emph{proof nets} to diagrammatically represent the composition of language meaning in Lambek calculus, \citet{pronounResolution} introduce novel string diagrams, where word meanings are represented as state vectors in Fock spaces, and copying mechanisms in language as \emph{projections} from these Fock spaces. We build on this work in the current article.

\citet{qnlpInPractice} train a machine learning model to predict whether a phrase contains a subject- or object-relative pronoun, e.g.:
\[\texttt{Device that detects planets} \qquad \text{or} \qquad \texttt{Device that observatory has.}\]
We demonstrate our approach to reasoning about ambiguity, using the same approach.





\subsection{Linguistics and Quantum Circuits}
\label{sec:linguisticsAndQCPaper}

The DisCoCat model allows sentences to be represented as diagrams. These diagrams can be mapped to quantum circuits, which encode the meanings of sentences as quantum states. 
In the current section, we briefly discuss the DisCoCat framework and its underlying category-theoretic origin. We continue by introducing several concepts from quantum computation, such as \emph{qubits} and \emph{quantum gates}. 
We explain how to establish a connection between these two frameworks, using work by \citet{qnlpInPractice} and \citet{coecke2020foundations}. 
We relate the presented theory to alternative approaches, such as the one by \citet{pronounResolution}, in which Fock spaces are used to model vector spaces in which words are represented.

Finally, we introduce the density matrix and discuss existing literature on it. We explain its application in QNLP and introduce the \emph{DisCoCirc} framework \citep{DisCoCirc}, in which sentences and words are modelled as density matrices. 

\subsubsection{The DisCoCat Framework}
\label{sec:DisCoCatPaper}


Language can be modelled in terms of its distributional and compositional properties. According to the distributional hypothesis, words that have related meanings tend to co-occur in similar contexts. Using the principle of compositionality, the meaning from sentences can be derived from its constituents depending on grammatical rules.
In the \emph{Categorical Compositional Distributional} (DisCoCat) framework \citep{discocat}, which combines these two properties of language, \textit{distributional} refers to the assignment of meaning to quantum states and \textit{compositional} refers to the grammatical derivations in the Pregroup formalism. 
\textit{Category Theory} combines these two aspects.

The compositions of tensors in the DisCoCat model are \textit{tensor networks}, which can be implemented using \textit{quantum circuits} \citep{tensorNetworksQML}. 

%
The \emph{pregroup grammar} \citep{pregroups}, a simplification of the \emph{Lambek calculus} \citep{orgLambek}, is the mathematical framework that guides the composition of quantum states that represent the word meanings in the DisCoCat framework.
Words are assigned atomic types $p$ corresponding to their grammatical function. 
The two reduction rules: 
\begin{equation}
    p^l \cdot p \rightarrow 1 \quad \qquad p \cdot p^r \rightarrow 1
\end{equation}
guide the composition of types of words in a sentence. 
A sentence is considered grammatical if it reduces to the canonical sentence-type $s$ upon multiplication of the types of its words:
\begin{equation}
    t_{\texttt{sentence}} = \prod_w t_w \rightarrow s 
\end{equation}
Consider, as an example, the sentence \texttt{Alice plays guitar}: assigning a type to each of the words (\texttt{Alice} $\rightarrow n$, \texttt{plays} $\rightarrow n^r \cdot s \cdot n^l$ and \texttt{guitar} $\rightarrow n$), following their grammatical function, and applying the reduction rules, we get:
\begin{equation}
    \texttt{Alice plays guitar}: \quad \quad n \cdot \big( n^r \cdot s \cdot n^l \big) \cdot n \rightarrow 1 \cdot s \cdot 1 \rightarrow s
\end{equation}
indicating that this sentence is indeed grammatical. 
The pregroup grammar can be identified as a \emph{compact closed category} of pregroups, \texttt{Preg}, to use it as a formalism guiding the composition of word meanings in a sentence. Compact closed categories have a diagrammatical language associated with them, called \textit{string diagrams}.
Grammar reductions as above can be presented in a diagram: 
\begin{equation}
    \begin{tikzpicture}
	\begin{pgfonlayer}{nodelayer}
		\node [style=none] (13) at (2.75, -0.25) {$n$};
		\node [style=none] (15) at (-2.75, -0.25) {$n$};
		\node [style=none] (16) at (-2.75, -1) {};
		\node [style=none] (17) at (-0.75, -1) {};
		\node [style=none] (23) at (-0.75, -0.25) {$n^r$};
		\node [style=none] (24) at (0, -0.25) {$s$};
		\node [style=none] (25) at (0.75, -0.25) {$n^l$};
		\node [style=none] (26) at (-2.75, 0.5) {\texttt{Alice}};
		\node [style=none] (27) at (2.75, 0.5) {\texttt{guitar}};
		\node [style=none] (28) at (0, 0.5) {\texttt{plays}};
		\node [style=none] (29) at (0.75, -1) {};
		\node [style=none] (30) at (2.75, -1) {};
		\node [style=none] (31) at (0, -1) {};
		\node [style=none] (32) at (0, -2.75) {};
	\end{pgfonlayer}
	\begin{pgfonlayer}{edgelayer}
		\draw [style=none, bend left=270, looseness=1.50] (16.center) to (17.center);
		\draw [style=none, bend left=270, looseness=1.50] (29.center) to (30.center);
		\draw [style=none] (31.center) to (32.center);
	\end{pgfonlayer}
\end{tikzpicture}
\end{equation}
Words are represented as differently ranked tensors, in the compact closed category of vector spaces and linear maps between them, \texttt{FVect} \citep{lambekVsLambek}. 
The DisCoCat framework is formally defined as the Cartesian product between \texttt{FVect} and \texttt{Preg}: DisCoCat is a category, whose objects are pairs of vector spaces $W$ and Pregroup types $p$:  $(W, p)$. 
The morphisms of the DisCoCat model are pairs of morphisms for $f$ a linear map and $[p \leq q]$ a Pregroup partial order, denoted:  
\begin{equation}
    (f: V \rightarrow W, [p \leq q])
\end{equation}
It is equipped with a tensor product given by the pointwise tensor of \texttt{Preg} and \texttt{FVect}, that is, $(V, p) \otimes (W, q) = (V \otimes W, p \cdot q)$, with a unit of $(\mathbb{R}, 1)$. 



The ranks of the resulting tensors correspond to the grammatical type assigned in \texttt{Preg}. 
Sticking with our example, \texttt{Alice} and \texttt{guitar} are assigned rank-one tensors (vectors) in the \textit{noun space}, labelled $N$: $\Vec{v}_{\texttt{Alice}}, \Vec{v}_{\texttt{guitar}} \in N$. 
And the transitive verb \texttt{plays} is a rank three tensor: $\Vec{v}_{\texttt{plays}} \in N \otimes S \otimes N$, where $S$ denotes the \textit{sentence space}. Thus, the types guide combination of the tensors to compose a single representation of the semantics of the sentence.

\begin{figure}
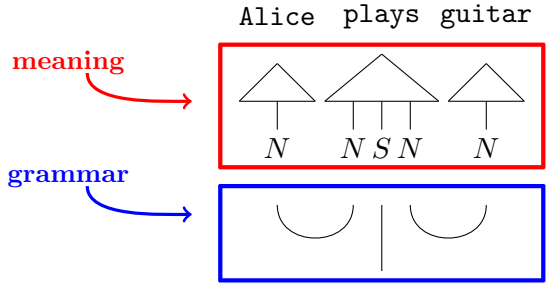

    \centering
    \ctikzfig{meaningGrammar}
    \caption{A DisCoCat diagram, composing tensors representing meanings of words, guided by grammar}
    \label{fig:DisCoCatModel}
\end{figure}

The composition of these tensors can be captured by string diagrams (\emph{DisCoCat diagrams}), where boxes represent tensors and wires represent how these tensors are composed (Figure \ref{fig:DisCoCatModel}). 
There is a substantial amount of literature on the DisCoCat model since it has been extended in many different directions. 
DisCoCat has been applied to translations between languages \citep{discocattranslations}, it has led to the development of \textit{language circuits} \citep{distilling}, and there are studies investigating the inner structure of words themselves, breaking them up into smaller parts \citep{grammarEq}. Furthermore, DisCoCat has been applied to music \citep{discocatmusic}, subsequently developing \textit{Quanthoven}, a quantum model to compose music. 

Among the concepts arising in category theory is the \textit{Frobenius Algebra}, a structure that arises in the category of finite dimensional vector spaces \texttt{FVect}, which allows information to be combined or deleted and is used in a QNLP context to deal with some function words that do not bear any contextual meaning, e.g., \texttt{that} or \texttt{and} \citep{relPronCoecke}.
Originally introduced by \citet{frobenius} in the context of group theory, their use in category theory is due to \citet{frobInCategory}. 
On a vector space level, this new structure introduces operations to expand vectors into matrices and to collapse matrices to vectors \citep{frobeniusCoecke}. 

Formally, a \textit{Frobenius algebra} is a tuple $(X, \Delta, \imath, \mu, \xi)$ in a symmetric monoidal category, where the following holds: 
\begin{equation}
    \begin{split}
        \Delta&: X \rightarrow X \otimes X \qquad \qquad \imath: X \rightarrow I \\
        \mu&: X \otimes X \rightarrow X \qquad \qquad \zeta: I \rightarrow X
    \end{split}
\end{equation}
The two morphisms $\Delta, \mu$,  satisfy a particular condition, called the \textit{Frobenius condition} \citep{frobeniusCoecke}.
Diagrammatically, these new morphisms can be associated with specific \emph{spiders} in the ZX-calculus \citep{zx1, zx2}. 
\begin{figure}
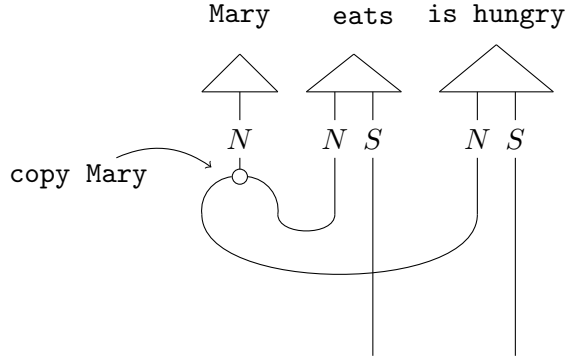

    \centering
    \ctikzfig{maryEatsAndDrinksSpiderExplanation}
    \caption{The meaning of the sentences \texttt{Mary eats. Mary is hungry.} as a pregroup diagram, after \citet{pronounResolution} who use a different framework based on projections from Fock space.}
    \label{fig:maryCopy}
\end{figure}
We can include these structures in existing DisCoCat diagrams. Consider the sentences: 
\begin{equation}
\label{eq:MaryEatsSheHungry}
    \texttt{Mary eats. She is hungry.}
\end{equation}
The word \texttt{Mary} is copied, because \texttt{Mary} both \texttt{eats} and \texttt{is hungry}, so that the sentence can be formulated as: 
\[\texttt{Mary eats. Mary is hungry.}\]

\citet{relPronCoecke} use Frobenius deletion maps to model subject- and object relative pronouns (\texttt{humans whom animals eat} vs. \texttt{humans who eat animals}). 

\citet{Wijnholds_2020} argues that the map $\Delta$ may be used to copy linguistic information, which gives the resulting DisCoCat diagram (Figure \ref{fig:maryCopy}).
In a different approach, \citet{pronounResolution} 
introduce a different mathematical tool to model the copying of linguistic information. 
They introduce a binary classification task, in which Variational Quantum Circuits (VQCs) are trained to predict whether a sentence contains a subject- or an object relative pronoun and model the copying process of linguistic information as \emph{projections} from \emph{truncated Fock spaces}. 
The Fock space for some vector space $V$ is \citet{fockspace}: 
\begin{equation}
\label{eq:FockSpace}
    F(V) = \bigoplus_{n=0}^\infty V^{\otimes n}
\end{equation}
where a \textit{truncated} Fock space then describes a Fock space restricted to its $k$-th tensor power  \citep{pronounResolution}: 
\begin{equation}
\label{eq:TruncatedFockSpace}
    F_{k}(V) = \bigoplus_{n=0}^{k} V^{\otimes n}
\end{equation}
The term \textit{projection} refers to accessing the $n$-th layer of the Fock space, which is an $n$-fold tensor product of the space $V$: 
\begin{equation}
\label{eq:projectionFockSpace}
    p_n: F_k(V) \rightarrow V^{\otimes n}
\end{equation}
In Section \ref{sec:linguisticsToQCPaper}, we relate the diagrammatic approach chosen by \citet{pronounResolution} to the pregroup approach and the arising quantum circuits. We thus argue why the above pregroup diagram indeed is correctly depicting the meaning of the above sentence.


The DisCoCat diagrams can be associated with \emph{tensor networks} \citep{qnlpInPractice, tensorNetworksQML}, which can themselves be associated with quantum circuits \citep{discocatAreTN}. 
This means that the meaning of a sentence can be represented as a quantum state on a quantum computer.

\subsubsection{Quantum Computing}
\label{sec:QuantumComputingPaper}


The current section focusses on parts of quantum computing that are relevant to the research presented in this paper. 
\citet{quantumComp}  give a thorough introduction to quantum computing in general.  


Quantum computing refers to the process of doing computations on quantum objects, such as \textit{qubits} (the quantum computational analogue of a \textit{bit}). The computing process can be described as a series of manipulations of the quantum states that these qubits are in. 
Quantum computers have been theorised to outperform classical computers on specific tasks,
for example, Grover's algorithm \citep{groversAlgorithm}, a quantum search algorithm, and Shor's algorithm \citep{shorsAlgorithm} for finding prime factors. 

Today, quantum computing finds itself in the \textit{Noisy Intermediate Scale Quantum} (NISQ) era, referring to the limited applicability of quantum computers due to high errors and small numbers of {qubits}. 
While in physics, quantum computers have been applied to various optimisation problems, such as the estimation of the ground state energy of a molecular Hamiltonian \citep{vqeMinEnergy}, the use of quantum computers is not limited to the realm of (molecular) physics. Its application in machine learning has led to the rise of a new field of research: \textit{Quantum Machine Learning} (QML), which is related to the field of Quantum Natural Language processing (QNLP) in that QML models are applied in QNLP to learn meanings of words and sentences.

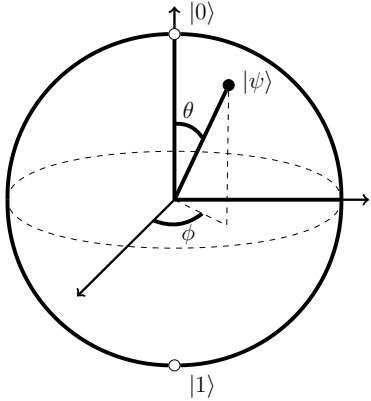
\begin{figure}
    \centering
    \resizebox{0.33\textwidth}{!}{\begin{tikzpicture}
	\begin{pgfonlayer}{nodelayer}
		\node [style=none] (0) at (0, 0) {};
		\node [style=white dot] (1) at (0, 6) {};
		\node [style=white dot] (2) at (0, -6) {};
		\node [style=none] (3) at (-6, 0) {};
		\node [style=none] (4) at (6, 0) {};
		\node [style=none] (5) at (5.23, 3) {};
		\node [style=none] (7) at (1.88, -0.93) {};
		\node [style=black dot] (8) at (1.95, 4.15) {};
		\node [font={\large}] (17) at (0.5, 3.25) {$\theta$};
		\node [font={\large}] (18) at (0.5, -1.25) {$\phi$};
		\node [style=none] (19) at (7, 0) {};
		\node [style=none] (20) at (-3.5, -3.5) {};
		\node [style=none] (21) at (0, 7) {};
		\node [style=none] (22) at (1, 2.25) {};
		\node [style=none] (23) at (0, 2.75) {};
		\node [style=none] (24) at (-0.75, -0.75) {};
		\node [style=none] (25) at (1, -0.5) {};
		\node [font={\large}] (26) at (3, 4.25) {$\ket{\psi}$};
		\node [font={\large}] (27) at (1, 6.75) {$\ket{0}$};
		\node [font={\large}] (28) at (1, -6.75) {$\ket{1}$};
	\end{pgfonlayer}
	\begin{pgfonlayer}{edgelayer}
		\draw [style=thick, bend left=45] (4.center) to (2);
		\draw [style=thick, bend left=45] (2) to (3.center);
		\draw [style=thick, bend left=45] (3.center) to (1);
		\draw [style=thick, bend left=45] (1) to (4.center);
		\draw [dashed, bend left=90, looseness=0.50] (4.center) to (3.center);
		\draw [dashed, bend left=90, looseness=0.50] (3.center) to (4.center);
		\draw [style=thick] (0.center) to (1);
		\draw [style=thick] (0.center) to (4.center);
		\draw [dashed] (0.center) to (7.center);
		\draw [style=thick] (0.center) to (8);
		\draw [style=thickdiredge] (1) to (21.center);
		\draw [style=thickdiredge] (4.center) to (19.center);
		\draw [style=thickdiredge] (0.center) to (20.center);
		\draw [dashed, style=none] (8) to (7.center);
		\draw [style=thick, bend left] (23.center) to (22.center);
		\draw [style=thick, bend right] (24.center) to (25.center);
	\end{pgfonlayer}
\end{tikzpicture}}
    \caption{The Bloch sphere visualisation of qubit state vectors, the black dot represents the state vector $\ket{\psi}$ of a qubit on the Bloch sphere}
    \label{fig:blochSphere}
\end{figure}

As for the relation between quantum computing and linguistics, the meaning of words and sentences are encoded onto one or more qubits. 
While a classical bit can only assume the values 0 and 1, the qubit naturally assumes values in between as well: 
\begin{equation}
\label{eq:qubit}
    \ket{{q}} = \alpha \ket{0} + \beta \ket{1}
\end{equation}
where $\ket{q}$ is the quantum state of a qubit, $\alpha, \beta \in \mathbb{C}$ are constants and $|\alpha|^2 + |\beta|^2 = 1$. 
Quantum gates are the basic building blocks of quantum circuits, which capture the algorithms to be executed on the qubits.
Mathematically, quantum gates are represented by unitary matrices $U$: 
\begin{equation}
    U^\dag U = I = U U^\dag
\end{equation}
where the symbol $\dag$ (\textit{dagger}) is the Hermitian adjoint.
Applying gate operations to qubits amounts to doing matrix multiplication of the unitary matrices with the states of the qubits, represented by vectors. 
The state vector of a qubit can be displayed on the \emph{Bloch sphere}, where the action of quantum gates on the qubit-states can be understood as rotations of the qubit state vectors on this sphere (Figure \ref{fig:blochSphere}). 
We give a brief overview of the most common gates. There are the Pauli-$X$, -$Y$, and -$Z$ gates which rotate the qubit state around the $x$-, $y$- or $z$-axis of the Bloch sphere, respectively, by an angle of $\pi$. These gates act on one qubit only and the matrix representations of the gates are the Pauli matrices: 
\begin{equation}
    X\, = \, \begin{pmatrix}
        0 & 1 \\ 1 & 0 
    \end{pmatrix} \qquad \quad
    Y\, = \, \begin{pmatrix}
        0 & -i \\ i & 0
    \end{pmatrix} \qquad \quad
     Z\, = \, \begin{pmatrix}
        1 & 0 \\ 0 & -1
    \end{pmatrix} 
\end{equation}
where the representation in quantum circuit notation is, respectively: 
\begin{equation}
    \begin{tikzpicture}
	\begin{pgfonlayer}{nodelayer}
		\node [style=hadamard, minimum width=7mm, minimum height=6mm] (29) at (-4.5, 0) { $X$};
		\node [style=none] (30) at (-6, 0) {};
		\node [style=none] (31) at (-3, 0) {};
		\node [style=hadamard, minimum width=7mm, minimum height=6mm] (32) at (1.5, 0) { $Y$};
		\node [style=none] (33) at (0, 0) {};
		\node [style=none] (34) at (3, 0) {};
		\node [style=hadamard, minimum width=7mm, minimum height=6mm] (35) at (7.5, 0) { $Z$};
		\node [style=none] (36) at (6, 0) {};
		\node [style=none] (37) at (9, 0) {};
	\end{pgfonlayer}
	\begin{pgfonlayer}{edgelayer}
		\draw (30.center) to (29);
		\draw (29) to (31.center);
		\draw (33.center) to (32);
		\draw (32) to (34.center);
		\draw (36.center) to (35);
		\draw (35) to (37.center);
	\end{pgfonlayer}
\end{tikzpicture}
\end{equation}
Note that in the quantum circuit notation, the qubits are usually represented by vertical lines, and the execution of gates happens from left to right in the circuit. 
Subsequently, generalisations of these gates, rotating around some given angle $\theta$ are introduced:
\begin{center}
\resizebox{\linewidth}{!}{
$
    R_X(\theta) \, = \, \begin{pmatrix}
        \cos(\theta / 2) & - i \sin(\theta / 2) \\ - i \sin(\theta / 2) & \cos(\theta / 2)
    \end{pmatrix} \quad 
    R_Y(\theta) \, = \, \begin{pmatrix}
        \cos(\theta / 2) & - \sin(\theta / 2) \\  \sin(\theta / 2) & \cos(\theta / 2)
    \end{pmatrix} \quad
    R_Z(\theta) \, = \, \begin{pmatrix}
        e^{-i \theta / 2} & 0 \\ 0 & e^{i \theta / 2}
    \end{pmatrix}
$
}
\end{center}
where the gates are represented in the circuit notation as: 
\begin{equation}
    \begin{tikzpicture}
	\begin{pgfonlayer}{nodelayer}
		\node [style=hadamard, minimum width=12mm, minimum height=8mm] (9) at (-10, 0) {\large $R_X(\theta)$};
		\node [style=none] (10) at (-12, 0) {};
		\node [style=none] (11) at (-8, 0) {};
		\node [style=none] (12) at (-2, 0) {};
		\node [style=none] (13) at (2, 0) {};
		\node [style=none] (14) at (8, 0) {};
		\node [style=none] (15) at (12, 0) {};
		\node [style=none] (17) at (0, 0) {};
		\node [style=none] (18) at (10, 0) {};
		\node [style=hadamard, minimum width=12mm, minimum height=8mm] (19) at (0, 0) {\large $R_Y(\theta)$};
		\node [style=hadamard, minimum width=12mm, minimum height=8mm] (20) at (10, 0) {\large $R_Z(\theta)$};
	\end{pgfonlayer}
	\begin{pgfonlayer}{edgelayer}
		\draw (17.center) to (12.center);
		\draw (17.center) to (13.center);
		\draw (14.center) to (18.center);
		\draw (18.center) to (15.center);
		\draw (10.center) to (9);
		\draw (9) to (11.center);
	\end{pgfonlayer}
\end{tikzpicture}
\end{equation}
One other very important single-qubit gate is the Hadamard gate \texttt{H}, represented by the  matrix 
\begin{equation}
\label{eq:h}
    \texttt{H} = \frac{1}{\sqrt{2}} \begin{pmatrix}
        1 & 1 \\ 1 & -1
    \end{pmatrix} \qquad \qquad \, \text{ and } \qquad \qquad \begin{tikzpicture}
	\begin{pgfonlayer}{nodelayer}
		\node [style=hadamard, minimum width=6mm, minimum height=6mm] (4) at (0, 0) {\large $H$};
		\node [style=none] (5) at (1.5, 0) {};
		\node [style=none] (6) at (-1.5, 0) {};
	\end{pgfonlayer}
	\begin{pgfonlayer}{edgelayer}
		\draw (6.center) to (4);
		\draw (4) to (5.center);
	\end{pgfonlayer}
\end{tikzpicture}

\end{equation}
Given a basis state $\ket{0}, \ket{1}$, the Hadamard gate maps it to an equal superposition state: 
\begin{equation*}
    \texttt{H} \ket{0} = \frac{1}{\sqrt{2}} \big(\ket{0} + \ket{1}\big) := \ket{+} \qquad \texttt{H} \ket{1} = \frac{1}{\sqrt{2}} \big(\ket{0} - \ket{1}\big) := \ket{-}
\end{equation*}
The Hadamard gate flips the $x-$ and $z-$axis, performing a basis change. 
Also of interest for our computations are gates that act on not only one but two or more qubits. 
An important example is the \emph{controlled} \texttt{NOT} (\texttt{CNOT}) gate with the following matrix representation: 
\begin{equation}
\label{eq:cnot}
    \texttt{CNOT} \, = \, \begin{pmatrix}
        1 & 0 & 0& 0 \\ 0 & 1 & 0 & 0 \\ 0 & 0& 0& 1 \\ 0 & 0& 1 & 0 
    \end{pmatrix}
\end{equation}
It is called {controlled} \texttt{NOT} gate because it performs a \texttt{NOT} operation on the second qubit only if the first qubit is in the state $\ket{1}$. If the first qubit is not in state $\ket{1}$, it leaves the second qubit unchanged. 
The \texttt{CNOT} gate is represented in circuit notation as:
\begin{equation}
    \begin{tikzpicture}
	\begin{pgfonlayer}{nodelayer}
		\node [style=vertex] (4) at (0, 1.5) {};
		\node [style=none] (5) at (0, 0) {$\bigoplus$};
		\node [style=none] (6) at (-1.5, 1.5) {};
		\node [style=none] (7) at (-1.5, 0) {};
		\node [style=none] (8) at (1.5, 0) {};
		\node [style=none] (9) at (1.5, 1.5) {};
	\end{pgfonlayer}
	\begin{pgfonlayer}{edgelayer}
		\draw (6.center) to (4);
		\draw (4) to (9.center);
		\draw (7.center) to (5.center);
		\draw (5.center) to (8.center);
		\draw (4) to (5.center);
	\end{pgfonlayer}
\end{tikzpicture}
\end{equation}
where in this notation, the black dot is the \textit{control} operation and the white dot with the cross is the \texttt{NOT} operation. 
Gates can act on more than two qubits as well. An example is the \emph{Toffoli} (\texttt{CCNOT}) gate,  
acting on the three-qubit states. The \texttt{CCNOT} gate is:
\begin{equation}
    \begin{tikzpicture}
	\begin{pgfonlayer}{nodelayer}
		\node [style=vertex] (0) at (0, 1) {};
		\node [style=none] (1) at (0, -0.5) {$\bigoplus$};
		\node [style=none] (2) at (-1.5, 1) {};
		\node [style=none] (3) at (-1.5, -0.5) {};
		\node [style=none] (4) at (1.5, -0.5) {};
		\node [style=none] (5) at (1.5, 1) {};
		\node [style=vertex] (6) at (0, 2.5) {};
		\node [style=none] (7) at (-1.5, 2.5) {};
		\node [style=none] (8) at (1.5, 2.5) {};
	\end{pgfonlayer}
	\begin{pgfonlayer}{edgelayer}
		\draw (2.center) to (0);
		\draw (0) to (5.center);
		\draw (3.center) to (1.center);
		\draw (1.center) to (4.center);
		\draw (0) to (1.center);
		\draw (7.center) to (6);
		\draw (6) to (8.center);
		\draw (6) to (0);
	\end{pgfonlayer}
\end{tikzpicture}
\end{equation}
Note that the set of Pauli $X$- and $Z$-, together with the \textit{phase}-, Hadamard- and \texttt{CNOT} gates generate the \textit{Clifford group}. Any gate in the Clifford group is called a Clifford gate. 
Clifford gates are of high interest, especially in the field of \textit{Quantum Error Correction}, due to the Gottesman-Knill theorem \citep{gottesmanPhD, gottesmanKnill}, which states that any quantum circuit that is made up only of Clifford gates can be simulated classically in polynomial time. Clifford gates have been used outside the realm of quantum error correction, e.g., to pre-optimise Variational Quantum Circuits \citep{cafqa}. 

\textit{Quantum circuits} are sequences of quantum gates and represent instructions on how to manipulate the qubits in the circuit. By convention, the $N_q$ qubits in a quantum circuit are initialised to the zero-state:
\begin{equation}
    \ket{\psi} = \underbrace{\ket{0} \otimes \ket{0} \otimes \ldots \otimes \ket{0}}_{ N_q \text{ times}} = \ket{00 \ldots 0}
\end{equation}
Quantum algorithms are carried out on these quantum circuits. 
In the context of this paper, quantum circuits are \emph{machine learning models} that are trained to predict sentence meaning. The training procedure corresponds to adjusting parameters of parameterised gates (similar to adjusting weights in a neural network training process). 

One last important quantum-theoretic concept is \emph{entanglement}. 
In quantum physics, a state is entangled, if it is composed of other states and these respective states cannot be described independently of the states of the others.
This phenomenon has no classical counterpart and is purely quantum theoretical. 

If a qubit that is entangled with other qubits is studied in isolation, its state is said to be \textit{mixed}. Upon measurement of one of the qubits constituting the entangled state, the individual qubits collapse to one possible measurement outcome, resulting in individual \textit{pure} states. 

More formally, consider two arbitrary quantum systems $A$ and $B$. 
These systems have Hilbert spaces $\mathcal{H}_{A}$ and $\mathcal{H}_{B}$ associated with them, so that the composed system's associated Hilbert space is the tensor product space $\mathcal{H}_A \otimes \mathcal{H}_B$. Consider the systems being in different quantum states $\ket{\psi}_A$ and $\ket{\psi}_B$.
The state of the composed system in the Hilbert space $\mathcal{H}_A \otimes \mathcal{H}_B$ is denoted $\ket{\psi}_{A \otimes B}$. 
This state is \emph{separable}, if the following holds:
\begin{equation}
    \ket{\psi}_{A \otimes B} = \ket{\psi}_A \otimes \ket{\psi}_B
\end{equation}
If the state is not separable, it is {entangled}.

In the history of quantum physics, the concept of entanglement has been met with scepticism. Albert Einstein, Boris Podolsky and Nathan Rosen, amongst others, considered such behaviour impossible, citing the so-called \textit{EPR Paradox} \citep{ERZParadox}. However, it turned out to not be a paradox after all, and quantum entanglement is an important part of quantum theory today. 
In particular, in the context of Quantum Natural Language Processing (QNLP), we will find the concept of entanglement to be very helpful in modeling the exact relation between quantum states representing the meanings of different words.

\subsubsection{From Linguistics to Quantum Circuits}
\label{sec:linguisticsToQCPaper}
\begin{figure}
    \centering
    \resizebox{0.55\textwidth}{!}{\begin{tikzpicture}
	\begin{pgfonlayer}{nodelayer}
		\node [style=none] (0) at (-3.75, 4.5) {};
		\node [style=none] (1) at (-2.75, 5.75) {};
		\node [style=none] (2) at (-1.75, 4.5) {};
		\node [style=none] (3) at (-2.75, 4.5) {};
		\node [style=none] (4) at (-2.75, 5) {{$0$}};
		\node [style=none] (5) at (-1, 4.5) {};
		\node [style=none] (6) at (0, 5.75) {};
		\node [style=none] (7) at (1, 4.5) {};
		\node [style=none] (8) at (0, 4.5) {};
		\node [style=none] (9) at (0, 5) {{$0$}};
		\node [style=black dot] (10) at (0, 1.25) {};
		\node [style=none] (11) at (1.75, 4.5) {};
		\node [style=none] (12) at (2.75, 5.75) {};
		\node [style=none] (13) at (3.75, 4.5) {};
		\node [style=none] (14) at (2.75, 4.5) {};
		\node [style=none] (15) at (2.75, 5) {{$0$}};
		\node [style=black dot] (16) at (2.75, -0.5) {};
		\node [style=none] (17) at (2.75, -0.5) {};
		\node [style=none] (18) at (0, -7.75) {};
		\node [style=none] (19) at (-3.5, -0.75) {};
		\node [style=none] (20) at (-3.5, -0.75) {};
		\node [style=none] (21) at (-4.75, 5.25) {};
		\node [style=none] (22) at (-4, 6.25) {};
		\node [style=none] (23) at (3.5, 6.25) {};
		\node [style=none] (24) at (4.25, 5.25) {};
		\node [style=none] (25) at (4.25, -0.25) {};
		\node [style=none] (26) at (2, -1.75) {};
		\node [style=none] (27) at (-2, -1.75) {};
		\node [style=none] (28) at (-4.75, 0.5) {};
		\node [style=hadamard, minimum width=6mm, minimum height=6mm] (29) at (-2.75, 3.25) {\large $H$};
		\node [style=hadamard, minimum width=6mm, minimum height=6mm] (30) at (0, 3.25) {\large $H$};
		\node [style=hadamard, minimum width=6mm, minimum height=6mm] (31) at (2.75, 3.25) {\large $H$};
		\node [style=hadamard, minimum width=6mm, minimum height=6mm] (32) at (-2.75, 1.25) {\large $R_Z(\theta_4)$};
		\node [style=hadamard, minimum width=6mm, minimum height=6mm] (33) at (0, -0.5) {\large $R_Z(\theta_5)$};
		\node [style=none] (34) at (7.5, 4.25) {};
		\node [style=none] (35) at (8.5, 4.25) {};
		\node [style=none] (36) at (7.5, 5.5) {};
		\node [style=none] (37) at (6.5, 4.25) {};
		\node [style=none] (38) at (7.5, 4.25) {};
		\node [style=none] (39) at (7.5, 4.75) {{$0$}};
		\node [style=none] (40) at (5.75, -2) {};
		\node [style=none] (41) at (9.25, -2) {};
		\node [style=none] (42) at (5.75, 6) {};
		\node [style=none] (43) at (9.25, 6) {};
		\node [style=hadamard, minimum width=6mm, minimum height=6mm] (44) at (7.5, 1) {\large $R_Z(\theta_7)$};
		\node [style=hadamard, minimum width=6mm, minimum height=6mm] (45) at (7.5, -1) {\large $R_X(\theta_8)$};
		\node [style=hadamard, minimum width=6mm, minimum height=6mm] (46) at (7.5, 3) {\large $R_X(\theta_6)$};
		\node [style=none] (47) at (0, 7) {\texttt{plays}};
		\node [style=none] (48) at (7.5, 7) {\texttt{guitar}};
		\node [style=none] (49) at (-8, 4.25) {};
		\node [style=none] (50) at (-7, 4.25) {};
		\node [style=none] (51) at (-8, 5.5) {};
		\node [style=none] (52) at (-9, 4.25) {};
		\node [style=none] (53) at (-8, 4.25) {};
		\node [style=none] (54) at (-8, 4.75) {{$0$}};
		\node [style=none] (55) at (-9.75, -2) {};
		\node [style=none] (56) at (-6.25, -2) {};
		\node [style=none] (57) at (-9.75, 6) {};
		\node [style=none] (58) at (-6.25, 6) {};
		\node [style=hadamard, minimum width=6mm, minimum height=6mm] (59) at (-8, 1) {\large $R_Z(\theta_2)$};
		\node [style=hadamard, minimum width=6mm, minimum height=6mm] (60) at (-8, -1) {\large $R_X(\theta_3)$};
		\node [style=hadamard, minimum width=6mm, minimum height=6mm] (61) at (-8, 3) {\large $R_X(\theta_1)$};
		\node [style=none] (62) at (-8, 7) {\texttt{Alice}};
		\node [style=none] (63) at (-3, -7.25) {};
		\node [style=hadamard, minimum width=6mm, minimum height=6mm] (64) at (-2.75, -7.25) {$\not \frown$};
		\node [style=hadamard, minimum width=6mm, minimum height=6mm] (65) at (-8, -7.25) {$\not \frown$};
		\node [style=vertex] (66) at (-8, -4.25) {};
		\node [style=none] (67) at (-2.75, -4.25) {$\bigoplus$};
		\node [style=none] (68) at (-8, -2.75) {};
		\node [style=none] (69) at (-2.75, -2.75) {};
		\node [style=none] (70) at (7.25, -7.25) {};
		\node [style=hadamard, minimum width=6mm, minimum height=6mm] (71) at (7.5, -7.25) {$\not \frown$};
		\node [style=hadamard, minimum width=6mm, minimum height=6mm] (72) at (2.75, -7.25) {$\not \frown$};
		\node [style=vertex] (73) at (2.75, -4.25) {};
		\node [style=none] (74) at (7.5, -4.25) {$\bigoplus$};
		\node [style=none] (75) at (2.75, -2.75) {};
		\node [style=none] (76) at (7.5, -2.75) {};
		\node [style=hadamard, minimum width=6mm, minimum height=6mm] (77) at (2.75, -5.5) {\large $H$};
		\node [style=hadamard, minimum width=6mm, minimum height=6mm] (78) at (-8, -5.5) {\large $H$};
	\end{pgfonlayer}
	\begin{pgfonlayer}{edgelayer}
		\draw (0.center) to (1.center);
		\draw (1.center) to (2.center);
		\draw (0.center) to (2.center);
		\draw (5.center) to (6.center);
		\draw (6.center) to (7.center);
		\draw (5.center) to (7.center);
		\draw (11.center) to (12.center);
		\draw (12.center) to (13.center);
		\draw (11.center) to (13.center);
		\draw [style=dashed] (28.center) to (21.center);
		\draw [style=dashed] (21.center) to (22.center);
		\draw [style=dashed] (22.center) to (23.center);
		\draw [style=dashed] (23.center) to (24.center);
		\draw [style=dashed] (24.center) to (25.center);
		\draw [style=dashed] (25.center) to (26.center);
		\draw [style=dashed] (26.center) to (27.center);
		\draw [style=dashed] (27.center) to (28.center);
		\draw (35.center) to (36.center);
		\draw (36.center) to (37.center);
		\draw (35.center) to (37.center);
		\draw [style=dashed] (42.center) to (40.center);
		\draw [style=dashed] (40.center) to (41.center);
		\draw [style=dashed] (41.center) to (43.center);
		\draw [style=dashed] (42.center) to (43.center);
		\draw (3.center) to (32);
		\draw (8.center) to (18.center);
		\draw (17.center) to (14.center);
		\draw (38.center) to (45);
		\draw (32) to (10);
		\draw (33) to (17.center);
		\draw (50.center) to (51.center);
		\draw (51.center) to (52.center);
		\draw (50.center) to (52.center);
		\draw [style=dashed] (57.center) to (55.center);
		\draw [style=dashed] (55.center) to (56.center);
		\draw [style=dashed] (56.center) to (58.center);
		\draw [style=dashed] (57.center) to (58.center);
		\draw (53.center) to (60);
		\draw (66) to (67.center);
		\draw (66) to (65);
		\draw (67.center) to (64);
		\draw (68.center) to (66);
		\draw (69.center) to (67.center);
		\draw (73) to (74.center);
		\draw (73) to (72);
		\draw (74.center) to (71);
		\draw (75.center) to (73);
		\draw (76.center) to (74.center);
		\draw (60) to (68.center);
		\draw (69.center) to (32);
		\draw (75.center) to (17.center);
		\draw (45) to (76.center);
	\end{pgfonlayer}
\end{tikzpicture}}
    \caption{Example circuit encoding the meaning of the sentence \texttt{Alice plays guitar}. Qubits are represented by vertical lines rather than horizontal lines in the usual quantum circuit notation, to emphasise the connection to DisCoCat diagrams. The combination of \texttt{Hadamard-}, \texttt{CNOT-}, and measurement gates correspond to the cup-shaped wires in DisCoCat diagrams.}
    \label{fig:explCircuit}
\end{figure}
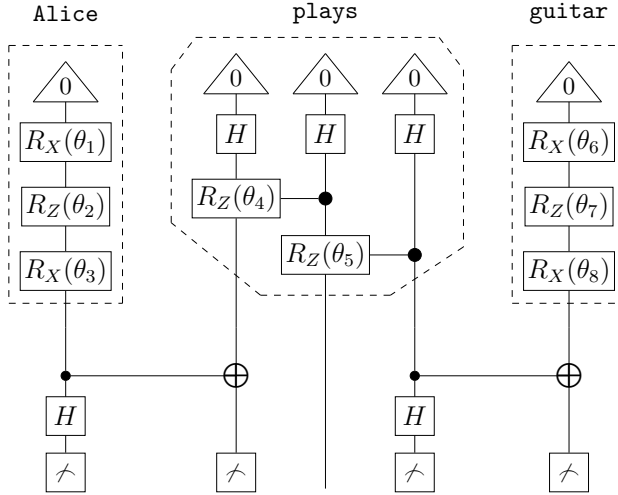

When the DisCoCat framework and quantum computing concepts are put together, quantum machine learning models arise: Variational Quantum Circuits (VQCs) that capture meanings of words and sentences as quantum (sub-)circuits. 
Following \citet{qnlpInPractice}, in quantum circuit representation, \texttt{Alice} can \texttt{play} her \texttt{guitar}, too (Figure \ref{fig:explCircuit}). 
Every word in the sentence is assigned a corresponding sub-circuit, and, in this case, one qubit encodes the meaning of the noun- and the sentence space, respectively. 
\citet{qnlpInPractice, coecke2020foundations} formalise this transition from DisCoCat to quantum circuits, which works via the idea of \emph{tensor networks} \citep{tensorNetworksQML}, and the field of \emph{Quantum Picturialism} \citep{coeckebook}. 
These fields arose from a particular notation introduced by \citet{penrose1971applications}, who presented a diagrammatical way of reasoning about tensors, in the context of quantum mechanics. 


Furthermore important is the choice of an \emph{ansatz}, which specifies how exactly individual qubits represent the meaning of words. One qubit might capture the meaning of a noun (making the complex noun-space two-dimensional), or two qubits might capture the meaning of a noun (making the complex noun-space four-dimensional), and the same holds for the sentence space. 
By choosing an ansatz, the particular combination of gates representing the meaning of some word is chosen, which influences the number of parameterised gates and how the gates entangle the qubit.
One may view this as the QNLP-equivalent of choosing the architecture of a neural network. 
In this work, the \texttt{IQP}-ansatz \citep{iqp} is chosen, and both the sentence- and the noun meaning are encoded onto one qubit, making sentence- and noun-space two-dimensional (Figure \ref{fig:explCircuit}). 



This article addresses a  binary classification task, which takes words in sentences as input data and outputs quantum states that represent the classification categories. One category will be represented by the state $\ket{1}$, and the second category by the state $\ket{0}$.

Note that there is a mechanism called \textit{amplitude encoding}\footnote{\url{https://pennylane.ai/qml/glossary/quantum_embedding/} (accessed 27.07.2024)}, which involves encoding classical data onto a quantum computer. When using amplitude encoding, some parameters of the model will be fixed, and not allowed to be learned in the training process. 
This is different to the approach of providing quantum states representing categories that the model is trained to predict.

In this work, the \emph{Binary Cross Entropy} loss function is used, of which the exact optimisation procedure of the parameters is explained by \citet{qnlpInPractice}. 
Obtaining the gradient for optimisation processes involving quantum circuits is a non-trivial task, because the  output of the quantum machine learning model is statistical.
The quantum computer does not output the probability distribution; the probability distribution needs to be reconstructed via numerous measurements.
This is why the gradient of Variational Quantum Circuits is approximated with an algorithm called \textit{Simultaneous Perturbation Stochastic Approximation} \citep[SPSA; ][]{spsa}, the state-of-the-art approach to approximating gradients in quantum machine learning models \citep{qnlpInPractice, qanswering, nearTermQC}.

\subsubsection{Copying linguistic Information}
\citet{pronounResolution} introduce a novel type of string diagram for their theory. In terms of these string diagrams, the projection from Fock space (Section \ref{sec:DisCoCatPaper}) is: 
\begin{equation}
    \label{eq:projectionDiagram}
    \begin{tikzpicture}
	\begin{pgfonlayer}{nodelayer}
		\node [style=element] (1) at (2.25, 1.5) {$v$};
		\node [style=new style 0] (2) at (2.25, -0.5) {$p_n$};
		\node [style=none] (3) at (0.75, -2.25) {};
		\node [style=none] (4) at (2.25, -2.25) {};
		\node [style=none] (5) at (2.925, -2) {$\cdots$};
		\node [style=none] (6) at (3.75, -2.25) {};
		\node [style=none] (12) at (2.75, 0.65) {$V$};
		\node [style=none] (14) at (4, -2) {$V$};
		\node [style=none] (18) at (0.55, -2) {$V$};
		\node [style=none] (19) at (1.8, -2) {$V$};
	\end{pgfonlayer}
	\begin{pgfonlayer}{edgelayer}
		\draw [style=downArrow, in=90, out=-135] (2) to (3.center);
		\draw [style=downArrow] (2) to (4.center);
		\draw [style=downArrow, in=90, out=-45] (2) to (6.center);
		\draw [style=thickArr] (1) to (2);
	\end{pgfonlayer}
\end{tikzpicture}
\end{equation}
In the following, we briefly relate this copying-approach to the approach of using Frobenius maps to copy linguistic information.  
\citet{pronounResolution} map the arising string diagrams to quantum circuits, following \citet{coecke2020foundations}. 
The arising quantum circuits are equivalent to those obtained from DisCoCat diagrams, with the exception of the way that the copying mechanism is modelled. 
In the DisCoCat framework, if a \texttt{noun} were to be copied, one obtains the following mapping:
\begin{equation}
    \begin{tikzpicture}
	\begin{pgfonlayer}{nodelayer}
		\node [style=none] (0) at (-3, 2) {};
		\node [style=none] (1) at (-2, 1) {};
		\node [style=none] (2) at (-4, 1) {};
		\node [style=none] (3) at (-3, 1) {};
		\node [style=none] (14) at (-3, 1) {};
		\node [style=none] (20) at (-3, 2.5) {\texttt{noun}};
		\node [style=white dot] (21) at (-3, 0) {};
		\node [style=none] (22) at (-4, -1) {};
		\node [style=none] (23) at (-2, -1) {};
		\node [style=none] (24) at (0.5, 0.5) {$\longmapsto$};
		\node [style=vertex] (26) at (4.5, -1) {};
		\node [style=none] (27) at (7, -1) {$\bigoplus$};
		\node [style=none] (28) at (7, 3) {\scriptsize $\ket{0}$};
		\node [style=none] (29) at (7, 2.5) {};
		\node [style=none] (30) at (4.5, -0.25) {};
		\node [style=none] (31) at (4.5, -1.75) {};
		\node [style=none] (32) at (7, -1.75) {};
		\node [style=hadamard, minimum width=6mm, minimum height=4mm] (33) at (4.5, -0.25) { $R_X(\theta^{\texttt{noun}}_3)$};
		\node [style=hadamard, minimum width=6mm, minimum height=4mm] (34) at (4.5, 0.75) { $R_Z(\theta^{\texttt{noun}}_2)$};
		\node [style=hadamard, minimum width=6mm, minimum height=4mm] (35) at (4.5, 1.75) {$R_X(\theta^{\texttt{noun}}_1)$};
		\node [style=none] (36) at (4.5, 3) {\scriptsize $\ket{0}$};
		\node [style=none] (37) at (4.5, 2.5) {};
		\node [style=none] (38) at (5.5, 2.5) {};
		\node [style=none] (39) at (3.5, 2.5) {};
		\node [style=none] (40) at (4.5, 4) {};
		\node [style=none] (41) at (6, 2.5) {};
		\node [style=none] (42) at (8, 2.5) {};
		\node [style=none] (43) at (7, 4) {};
		\node [style=none] (44) at (-4, -1.5) {$N$};
		\node [style=none] (45) at (-2, -1.5) {$N$};
	\end{pgfonlayer}
	\begin{pgfonlayer}{edgelayer}
		\draw [style=none] (2.center) to (1.center);
		\draw [style=none] (1.center) to (0.center);
		\draw [style=none] (0.center) to (2.center);
		\draw (14.center) to (21);
		\draw [bend left=45] (22.center) to (21);
		\draw [bend left=45] (21) to (23.center);
		\draw (26) to (27.center);
		\draw (27.center) to (29.center);
		\draw (26) to (30.center);
		\draw (31.center) to (26);
		\draw (32.center) to (27.center);
		\draw (33) to (34);
		\draw (34) to (35);
		\draw (35) to (37.center);
		\draw (39.center) to (40.center);
		\draw (40.center) to (38.center);
		\draw (41.center) to (43.center);
		\draw (43.center) to (42.center);
		\draw (39.center) to (38.center);
		\draw (41.center) to (42.center);
	\end{pgfonlayer}
\end{tikzpicture}
\end{equation}
Formally, the above-mentioned spider and the CNOT gate are equivalent:
\begin{equation}
\label{eq:spiderCnot}
    \begin{tikzpicture}
	\begin{pgfonlayer}{nodelayer}
		\node [style=white dot] (0) at (-4, 0) {};
		\node [style=none] (1) at (-5, -1) {};
		\node [style=none] (2) at (-3, -1) {};
		\node [style=none] (3) at (-4, 1) {};
		\node [style=none] (4) at (-1, 0) {$=$};
		\node [style=vertex] (5) at (1, -0.25) {};
		\node [style=none] (6) at (2.75, -0.25) {$\bigoplus$};
		\node [style=none] (7) at (2.75, 1.25) {\scriptsize $\ket{0}$};
		\node [style=none] (8) at (2.75, 0.75) {};
		\node [style=none] (9) at (1, 1.25) {};
		\node [style=none] (10) at (1, -1) {};
		\node [style=none] (11) at (2.75, -1) {};
	\end{pgfonlayer}
	\begin{pgfonlayer}{edgelayer}
		\draw [bend left=45] (1.center) to (0);
		\draw [bend left=45] (0) to (2.center);
		\draw (0) to (3.center);
		\draw (5) to (6.center);
		\draw (6.center) to (8.center);
		\draw (5) to (9.center);
		\draw (10.center) to (5);
		\draw (11.center) to (6.center);
	\end{pgfonlayer}
\end{tikzpicture}
\end{equation}
In contrast, \citet{pronounResolution} choose the following mapping of a second layer projection from Fock space:
\begin{equation}
    \begin{tikzpicture}
	\begin{pgfonlayer}{nodelayer}
		\node [style=none] (0) at (-3.5, 0) {$=$};
		\node [style=none] (8) at (-1.25, -0.75) {};
		\node [style=none] (9) at (-0.25, -0.75) {};
		\node [style=none] (10) at (3.25, 0) {$\longmapsto$};
		\node [style=vertex] (11) at (7, -0.5) {};
		\node [style=none] (13) at (9.5, 2) {\scriptsize $\ket{0}$};
		\node [style=none] (14) at (9.5, 1.5) {};
		\node [style=none] (16) at (7, -1.5) {};
		\node [style=none] (17) at (9.5, -1.5) {};
		\node [style=hadamard, minimum width=4mm, minimum height=4mm] (20) at (7, 0.75) {$H$};
		\node [style=none] (21) at (7, 2) {\scriptsize $\ket{0}$};
		\node [style=none] (22) at (7, 1.5) {};
		\node [style=none] (23) at (8, 1.5) {};
		\node [style=none] (24) at (6, 1.5) {};
		\node [style=none] (25) at (7, 3) {};
		\node [style=none] (26) at (8.5, 1.5) {};
		\node [style=none] (27) at (10.5, 1.5) {};
		\node [style=none] (28) at (9.5, 3) {};
		\node [style=hadamard, minimum width=4mm, minimum height=4mm] (29) at (9.5, 0.75) {$H$};
		\node [style=hadamard, minimum width=4mm, minimum height=4mm] (30) at (9.5, -0.5) { $R_Z(\theta^{\texttt{noun}}_1)$};
		\node [style=element] (31) at (-6.55, 2) {\texttt{noun}};
		\node [style=new style 0] (32) at (-6.55, 0) {$p_2$};
		\node [style=none] (33) at (-8.05, -1.75) {};
		\node [style=none] (36) at (-5.05, -1.75) {};
		\node [style=none] (37) at (-8.75, -1.5) {$N$};
		\node [style=none] (38) at (-4.5, -1.5) {$N$};
		\node [style=none] (39) at (-1.25, -1.25) {$N$};
		\node [style=none] (40) at (-0.25, -1.25) {$N$};
		\node [style=element] (41) at (-0.75, 1.25) {\texttt{noun}};
		\node [style=none] (42) at (-1.25, 1) {};
		\node [style=none] (43) at (-0.25, 1) {};
	\end{pgfonlayer}
	\begin{pgfonlayer}{edgelayer}
		\draw (16.center) to (11);
		\draw (20) to (22.center);
		\draw (24.center) to (25.center);
		\draw (25.center) to (23.center);
		\draw (26.center) to (28.center);
		\draw (28.center) to (27.center);
		\draw (24.center) to (23.center);
		\draw (26.center) to (27.center);
		\draw (11) to (20);
		\draw (11) to (30);
		\draw (30) to (17.center);
		\draw (30) to (14.center);
		\draw [style=downArrow, in=90, out=-135] (32) to (33.center);
		\draw [style=downArrow, in=90, out=-45] (32) to (36.center);
		\draw [style=thickArr] (31) to (32);
		\draw (8.center) to (42.center);
		\draw (43.center) to (9.center);
	\end{pgfonlayer}
\end{tikzpicture}
\end{equation}
They choose to model the noun as consisting of two qubits, subsequently following the IQP-ansatz for two qubits to encode the meaning of the copied noun, rather than modelling the noun on one qubit which is copied onto a second qubit, as in the pregroup approach.
In this approach, the noun only exists in a \enquote{copy-state} representing two nouns at once, rather than the noun being represented as two individual nouns that are the same. 
\citet{pronounResolution} model sentences with subject- and object-relative pronouns (Example \ref{eq:MaryEatsSheHungry}), using projections from Fock space (Equation \ref{eq:projectionDiagram}) rather than the Frobenius copying map (Figure \ref{fig:maryCopy}). 






\subsubsection{Density Matrices}
\label{sec:densityMatricesPaper}

Density matrices are of central interest in the context of QNLP due to their ability to, on the one hand, model ambiguity in language and, on the other hand, capture the hierarchy between word meanings. In the current work, we are interested in the former aspect. 

The density matrix is a positive semi-definite, hermitian operator $\rho$ with trace one that represents a probability distribution over quantum states. Quantum states represented by the density matrix are called \emph{mixed states}, while the state vector can only represent \textit{pure states}. 
A density matrix $\rho$ is generally defined as: 
\begin{equation}
    \rho = \sum_i p_i \ket{\psi_i} \bra{\psi_i}
\end{equation}
for a series of pure quantum states $\ket{\psi_1}, \ket{\psi_2}, \ldots$ and corresponding probabilities $p_1, p_2, \ldots$, that sum to one. Usually, the density matrix is used to describe the state of a system which is \emph{entangled} with another quantum system, or in a case where there is information missing in the system (e.g., about the initialisation of quantum states). 
The maps acting on density matrices, preserving their positive semi-definiteness are \emph{completely positive maps}.

In quantum physics, an \emph{observable} is a quantity that can be measured. It is represented mathematically by linear operators. 
Consider some observable $A$, in the system described by $\rho$. We obtain the expected value of $A$ in this system as: 
\begin{equation}
    \langle A \rangle = \Tr \big( \rho \, A \big)
\end{equation}
where, in contrast, the expectation value for the state vector ($\ket{\psi}$) case is described by: 
\begin{equation}
    \langle A \rangle = \braket{ \, \psi \, | \, A \, | \, \psi \, }
\end{equation} 
where $\ket{\psi}$ is some pure state. 
According to the Schrödinger-HJW theorem\footnote{The Schrödinger-HJW theorem can be seen as a special case of the \emph{Stinespring Dilation} \citep{coeckebook}} \citep{shjw}, a mixed state $\rho$ can be \textit{purified}: it can be represented by a \emph{partial trace} of a \textit{pure state} in a composite Hilbert space $\mathcal{H} = \mathcal{H}_1 \otimes \mathcal{H}_2$. 
This means that there always exists a pure state $\psi_{12}$, for which: 
\begin{equation}
    \rho = \Tr_2 \big( \ket{\psi_{12}} \bra{\psi_{12}} \big)
\end{equation}
where $\Tr_2$ is the \textit{partial trace} over Hilbert space $\mathcal{H}_2$. This density matrix is called a \textit{reduced density matrix}. 

In Section \ref{sec:idea}, mixed states are explicitly constructed from pure states to model density matrices on quantum circuits. 
This process is related to the \emph{discarding} effect. \citet{coeckebook} define the process of \textit{discarding} a qubit as tracing out its corresponding Hilbert space.


On a quantum circuit level, the process of discarding a qubit amounts to not measuring it followed by ignoring the qubit. 
\begin{figure}
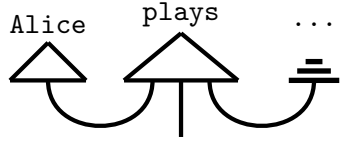

    \centering
    \ctikzfig{alicePlaysBlank}
    \caption{The diagram encoding the meaning of the sentence \texttt{Alice plays \ldots}, where the three dots indicate that this word is not available}
    \label{fig:alicePlaysBlank}
\end{figure}
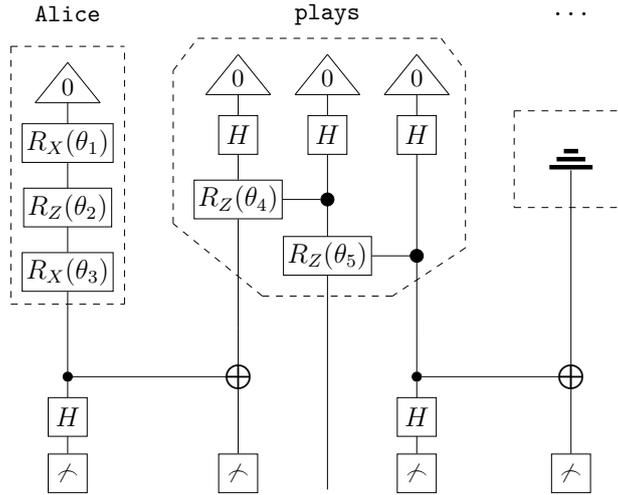
\begin{figure}
    \centering
    \resizebox{0.55\textwidth}{!}{\begin{tikzpicture}
	\begin{pgfonlayer}{nodelayer}
		\node [style=none] (0) at (-3.75, 4.25) {};
		\node [style=none] (1) at (-2.75, 5.5) {};
		\node [style=none] (2) at (-1.75, 4.25) {};
		\node [style=none] (3) at (-2.75, 4.25) {};
		\node [style=none] (4) at (-2.75, 4.75) {{$0$}};
		\node [style=none] (5) at (-1, 4.25) {};
		\node [style=none] (6) at (0, 5.5) {};
		\node [style=none] (7) at (1, 4.25) {};
		\node [style=none] (8) at (0, 4.25) {};
		\node [style=none] (9) at (0, 4.75) {{$0$}};
		\node [style=black dot] (10) at (0, 1) {};
		\node [style=none] (11) at (1.75, 4.25) {};
		\node [style=none] (12) at (2.75, 5.5) {};
		\node [style=none] (13) at (3.75, 4.25) {};
		\node [style=none] (14) at (2.75, 4.25) {};
		\node [style=none] (15) at (2.75, 4.75) {{$0$}};
		\node [style=black dot] (16) at (2.75, -0.75) {};
		\node [style=none] (17) at (2.75, -0.75) {};
		\node [style=none] (18) at (0, -8) {};
		\node [style=none] (19) at (-3.5, -1) {};
		\node [style=none] (20) at (-3.5, -1) {};
		\node [style=none] (21) at (-4.75, 5) {};
		\node [style=none] (22) at (-4, 6) {};
		\node [style=none] (23) at (3.5, 6) {};
		\node [style=none] (24) at (4.25, 5) {};
		\node [style=none] (25) at (4.25, -0.5) {};
		\node [style=none] (26) at (2, -2) {};
		\node [style=none] (27) at (-2, -2) {};
		\node [style=none] (28) at (-4.75, 0.25) {};
		\node [style=hadamard, minimum width=6mm, minimum height=6mm] (29) at (-2.75, 3) {\large $H$};
		\node [style=hadamard, minimum width=6mm, minimum height=6mm] (30) at (0, 3) {\large $H$};
		\node [style=hadamard, minimum width=6mm, minimum height=6mm] (31) at (2.75, 3) {\large $H$};
		\node [style=hadamard, minimum width=6mm, minimum height=6mm] (32) at (-2.75, 1) {\large $R_Z(\theta_4)$};
		\node [style=hadamard, minimum width=6mm, minimum height=6mm] (33) at (0, -0.75) {\large $R_Z(\theta_5)$};
		\node [style=none] (40) at (5.75, 0.75) {};
		\node [style=none] (41) at (9.25, 0.75) {};
		\node [style=none] (42) at (5.75, 3.75) {};
		\node [style=none] (43) at (9.25, 3.75) {};
		\node [style=none] (44) at (0, 6.75) {\texttt{plays}};
		\node [style=none] (45) at (7.5, 6.75) {\texttt{...}};
		\node [style=none] (46) at (-8, 4) {};
		\node [style=none] (47) at (-7, 4) {};
		\node [style=none] (48) at (-8, 5.25) {};
		\node [style=none] (49) at (-9, 4) {};
		\node [style=none] (50) at (-8, 4) {};
		\node [style=none] (51) at (-8, 4.5) {{$0$}};
		\node [style=none] (52) at (-9.75, -2.25) {};
		\node [style=none] (53) at (-6.25, -2.25) {};
		\node [style=none] (54) at (-9.75, 5.75) {};
		\node [style=none] (55) at (-6.25, 5.75) {};
		\node [style=hadamard, minimum width=6mm, minimum height=6mm] (56) at (-8, 0.75) {\large $R_Z(\theta_2)$};
		\node [style=hadamard, minimum width=6mm, minimum height=6mm] (57) at (-8, -1.25) {\large $R_X(\theta_3)$};
		\node [style=hadamard, minimum width=6mm, minimum height=6mm] (58) at (-8, 2.75) {\large $R_X(\theta_1)$};
		\node [style=none] (59) at (-8, 6.75) {\texttt{Alice}};
		\node [style=none] (60) at (-3, -7.5) {};
		\node [style=hadamard, minimum width=6mm, minimum height=6mm] (61) at (-2.75, -7.5) {$\not \frown$};
		\node [style=hadamard, minimum width=6mm, minimum height=6mm] (62) at (-8, -7.5) {$\not \frown$};
		\node [style=vertex] (63) at (-8, -4.5) {};
		\node [style=none] (64) at (-2.75, -4.5) {$\bigoplus$};
		\node [style=none] (65) at (-8, -3) {};
		\node [style=none] (66) at (-2.75, -3) {};
		\node [style=none] (67) at (7.25, -7.5) {};
		\node [style=hadamard, minimum width=6mm, minimum height=6mm] (68) at (7.5, -7.5) {$\not \frown$};
		\node [style=hadamard, minimum width=6mm, minimum height=6mm] (69) at (2.75, -7.5) {$\not \frown$};
		\node [style=vertex] (70) at (2.75, -4.5) {};
		\node [style=none] (71) at (7.5, -4.5) {$\bigoplus$};
		\node [style=none] (72) at (2.75, -3) {};
		\node [style=hadamard, minimum width=6mm, minimum height=6mm] (74) at (2.75, -5.75) {\large $H$};
		\node [style=hadamard, minimum width=6mm, minimum height=6mm] (75) at (-8, -5.75) {\large $H$};
		\node [style=upground] (87) at (7.5, 2.25) {};
	\end{pgfonlayer}
	\begin{pgfonlayer}{edgelayer}
		\draw (0.center) to (1.center);
		\draw (1.center) to (2.center);
		\draw (0.center) to (2.center);
		\draw (5.center) to (6.center);
		\draw (6.center) to (7.center);
		\draw (5.center) to (7.center);
		\draw (11.center) to (12.center);
		\draw (12.center) to (13.center);
		\draw (11.center) to (13.center);
		\draw [style=dashed] (28.center) to (21.center);
		\draw [style=dashed] (21.center) to (22.center);
		\draw [style=dashed] (22.center) to (23.center);
		\draw [style=dashed] (23.center) to (24.center);
		\draw [style=dashed] (24.center) to (25.center);
		\draw [style=dashed] (25.center) to (26.center);
		\draw [style=dashed] (26.center) to (27.center);
		\draw [style=dashed] (27.center) to (28.center);
		\draw [style=dashed] (42.center) to (40.center);
		\draw [style=dashed] (40.center) to (41.center);
		\draw [style=dashed] (41.center) to (43.center);
		\draw [style=dashed] (42.center) to (43.center);
		\draw (3.center) to (32);
		\draw (8.center) to (18.center);
		\draw (17.center) to (14.center);
		\draw (32) to (10);
		\draw (33) to (17.center);
		\draw (47.center) to (48.center);
		\draw (48.center) to (49.center);
		\draw (47.center) to (49.center);
		\draw [style=dashed] (54.center) to (52.center);
		\draw [style=dashed] (52.center) to (53.center);
		\draw [style=dashed] (53.center) to (55.center);
		\draw [style=dashed] (54.center) to (55.center);
		\draw (50.center) to (57);
		\draw (63) to (64.center);
		\draw (63) to (62);
		\draw (64.center) to (61);
		\draw (65.center) to (63);
		\draw (66.center) to (64.center);
		\draw (70) to (71.center);
		\draw (70) to (69);
		\draw (71.center) to (68);
		\draw (72.center) to (70);
		\draw (57) to (65.center);
		\draw (66.center) to (32);
		\draw (72.center) to (17.center);
		\draw (71.center) to (87);
	\end{pgfonlayer}
\end{tikzpicture}}
    \caption{The quantum circuit encoding the meaning of the sentence \texttt{Alice plays \ldots}, where the three dots indicate that the respective word is missing from the sentence}
    \label{fig:alicePlaysBlankQCTwo}
\end{figure}
The diagrammatic depiction of the discarding map:
\begin{equation}
\label{eq:discarding}
    \begin{tikzpicture}
	\begin{pgfonlayer}{nodelayer}
		\node [style=upground] (0) at (0, 2) {};
		\node [style=none] (1) at (0, 1) {};
	\end{pgfonlayer}
	\begin{pgfonlayer}{edgelayer}
		\draw [style=doubled] (1.center) to (0);
	\end{pgfonlayer}
\end{tikzpicture}

\end{equation}
could be used in a composite diagram (Figure \ref{fig:alicePlaysBlank}), corresponding to a quantum circuit (Figure \ref{fig:alicePlaysBlankQCTwo}).

\textit{Von Neumann entropy} is a measure to quantify the mixedness of a quantum state $\rho$.
It can be considered the quantum theoretical equivalent to the \textit{Shannon entropy} \citep{shannon1948mathematical}. In information theory, the {Shannon entropy} of some variable is a measure of \textit{uncertainty} or average \textit{information-content} in this variable:
\begin{equation}
    \label{eq:shannon}
    H_\text{Shannon} = - \sum_{x \in \mathcal{X}} p(x) \log_2 p(x)
\end{equation}
where $x$ are the values of a random variable in the set $\mathcal{X}$ and $p(x) \in [0,1]$. 
The Von Neumann entropy \citep{vonNeumann} is defined as: 
\begin{equation}
\label{eq:vonNeumannEntropy}
    S_\text{Von Neumann} = - \Tr \big( \, \rho \ln \rho  \, \big)
\end{equation}
where $\rho$ is a density matrix describing a quantum physical system and $\Tr$ is the \textit{trace}. 
The Von Neumann entropy assumes values between 0 (for a pure state) and $\ln (d)$ (for a completely mixed state) for $d$ the dimension of the corresponding Hilbert space. 
In this work, the Von Neumann entropy is used to reason about the information and uncertainty captured in certain quantum circuits representing words and sentences. 
In addition, we use the \emph{fidelity} as a measure of the closeness of two density matrices. 
The {fidelity} of two density matrices $\rho, \sigma$ is defined as: 
\begin{equation}
\label{eq:fidelity}
    F(\rho, \sigma) = \Tr \Big (\sqrt{\sqrt{\rho} \sigma \sqrt{\rho}}\Big)
\end{equation}
and \citet{densMat} argue that it is a suitable metric for the comparison of density matrices.
In the context of language, density matrices are applied to model ambiguity and hierarchic relations between words \citep{coecke2020foundations}. 
Consider the example of the word \verb|bank|. This could be a \texttt{financial institute}, it could be a \texttt{river bank}, or it could be a \texttt{computer memory bank}. There are several meanings for the same word, which makes it \textit{ambiguous}. To capture the {ambiguity} in this word, one can use the density matrix as a probability distribution over the pure states capturing the meanings of the words:
\begin{equation}
    \begin{split}
        \rho_{\texttt{bank}} &= \alpha \, \ket{\texttt{bank}_{\texttt{river}}} \bra{\texttt{bank}_{\texttt{river}}} \\
                    &+ \beta \, \ket{\texttt{bank}_{\texttt{finance}}} \bra{\texttt{bank}_{\texttt{finance}}} \\
                    &+ \gamma \,\ket{\texttt{bank}_{\texttt{memory}}} \bra{\texttt{bank}_{\texttt{memory}}}
    \end{split}
\end{equation}
where $\alpha, \beta, \gamma$ are positive, real numbers that sum to one. 
Words in context, represented by other quantum states, disambiguate the word represented by the density matrix. 
On a lower level, all the pure states $\ket{\texttt{bank}_{\texttt{river}}}, \ket{\texttt{bank}_{\texttt{finance}}}, \ldots$ are represented in terms of some basis states. 
If a word similar to the word \verb|riverbank| appears in context, like \verb|fish|, this disambiguates the word \verb|bank| modelled by the density matrix $\rho_{\texttt{bank}}$ above. 

A density matrices is a series of weighted projection operators, which is an intuitive approach to the above. By composing the density matrices $\rho_{\texttt{fish}}$ and $\rho_{\texttt{bank}}$, the \texttt{fish} part is projected out from $\rho_{\texttt{bank}}$, which is \texttt{riverbank}, assuming that the state vectors corresponding to the three terms in the density matrix are orthogonal to each other. 
The pure eigenstates of the density matrix (\texttt{riverbank, financial bank, computer memory bank}) are usually themselves made up of more basic words. It is common to train high dimensional vector spaces with order \num{1000} basis words; these vector spaces are called \textit{count-based} vector spaces. They are trained based on \textit{context-windows} around the word of interest \citep{word2vecSkipGram, word2vecContBagOfWords}.

The diagrammatic effect that corresponds to moving from state vectors to density matrices representing word meanings, is formally introduced by \citet{doublingPaper} as \textit{doubling}. This means that diagrams in which word meanings are represented as density matrices are drawn with \textit{thick} wires. 

As such, density matrices may be used to model probability distributions over \emph{word meanings}, and as a result of plugging in these density matrices in the sentence, over \emph{sentence meanings}. 
However, the probability distribution over the different sentence meanings is a probability distribution over sentences that have the same syntax. 
This does not involve creating probability distributions over sentences containing different syntax. 
In this work, we illustrate the advantages of using probability distributions over quantum processes for this task as well and argue that this allows for more general, human-like learning of machines. 
 
\begin{figure}
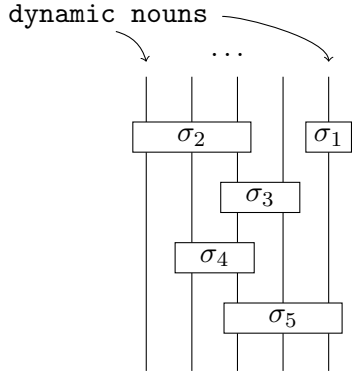

    \centering
    \ctikzfig{sentences}
    \caption{The action of sentences $\sigma_1, \ldots \sigma_5$ on {dynamic nouns} in the DisCoCirc framework}
    \label{fig:sentences}
\end{figure}

\begin{figure}
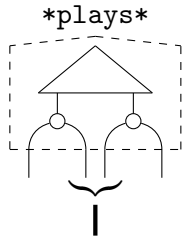

    \centering
    \ctikzfig{verb}
    \caption{The compact representation of the verb \texttt{plays}. 
    }
    \label{fig:verb}
\end{figure}

Rather than modelling the interaction of words to represent the meaning of sentences, as in the DisCoCat framework, the DisCoCirc framework models the interaction of sentences with each other. 
Following \citet{DisCoCircBook}, text can be seen as \textit{a process that alters the meaning of words}.
The diagram one ends up with when investigating how sentences $\sigma_1, \ldots, \sigma_5$ act on a number of words is depicted in Figure \ref{fig:sentences} (after \citet{DisCoCirc}). 
The words that are acted upon by sentences are only nouns, never verbs or adjectives.
Some nouns are termed \textit{dynamic nouns} whose meanings are altered and updated as the text proceeds. 
The updating mechanism with which this happens is discussed further below, and a detailed overview is given by \citet{meaningUpdatingDensMat}. 
Nouns that are not dynamic, or adjectives and verbs, are called \textit{static}, which means that their meaning is not updated by other words. 
To model updating of language meaning, both dynamic and static words are modelled using density matrices. 
In general, \citet{DisCoCirc} argues that the identification of dynamic nouns is non-trivial and requires additional research. 
One may choose a \emph{compact} representation \citep{compactVerbOne, compactVerbTwo} of a verb (Figure \ref{fig:verb}),
where the verb $^*\texttt{plays}^*$ lives in a lower dimensional vector space than the word \texttt{plays}. 
Here, the sentence type is the composition of two noun types, which means that the sentence space is the tensor product space $N \otimes N$. Note that this does not correspond to doubling the wires, and rather to a different way of modelling verbs as tensors. 

\begin{figure}
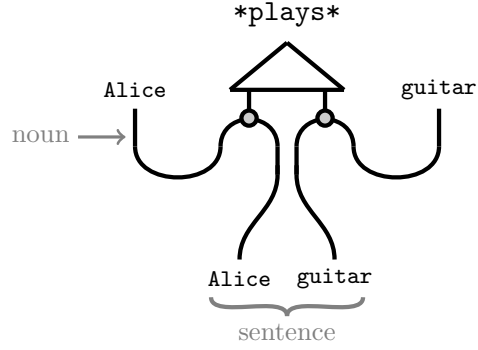

    \centering
    \ctikzfig{alicePlaysGuitarCompactDisCoCirc}
    \caption{\texttt{Alice} and \texttt{guitar} are density matrices whose meanings are updated by the verb \texttt{plays}, which entangles the nouns. The grey dot indicates the operation composing density matrices and is either the {fuzz} (Equation \ref{eq:fuzz}) or the {phaser} (Equation \ref{eq:phaser})}
    \label{fig:bobBitesAlice}
\end{figure}

Following \citet{meaningUpdatingDensMat}, words can be modelled as density matrices updating each other (Figure \ref{fig:bobBitesAlice}),
where \texttt{Bob} and \texttt{Alice} are dynamic nouns, and the sentence is the simple action of the verb \texttt{plays} on the nouns, indicating that \texttt{Alice} is playing the \texttt{guitar}, which, in turn, is being played by \texttt{Alice}. Note the \textit{thick} wires, indicating the use of density matrices. Furthermore, \citet{meaningUpdatingDensMat} introduce a grey dot as the symbol indicating the composition of density matrices. 

Updating one word using another word, where both words are represented by density matrices (consider, e.g., that \texttt{Alice} is now \texttt{sad}), is:
\begin{equation}
    \begin{tikzpicture}[scale=0.80]
	\begin{pgfonlayer}{nodelayer}
		\node [style=none] (0) at (-0.5, -2) {};
		\node [style=none] (1) at (-0.5, 1.75) {};
		\node [style=none, boldedge] (3) at (1, 0.25) {};
		\node [style=gray dot, boldedge] (4) at (-0.5, -1.25) {};
		\node [style=none] (5) at (1, 1.25) {};
		\node [style=none] (6) at (2, 0.25) {};
		\node [style=none] (7) at (0, 0.25) {};
		\node [style=none] (8) at (1, 0.25) {};
		\node [style=none] (9) at (-0.5, 2.75) {};
		\node [style=none] (10) at (0.5, 1.75) {};
		\node [style=none] (11) at (-1.5, 1.75) {};
		\node [style=none] (12) at (-0.5, 1.75) {};
		\node [style=none] (13) at (1.75, 1.75) {\texttt{sad}};
		\node [style=none] (14) at (-0.5, 3.25) {\texttt{Alice}};
	\end{pgfonlayer}
	\begin{pgfonlayer}{edgelayer}
		\draw [style=boldedge, in=45, out=-90] (3.center) to (4);
		\draw [style=boldedge] (4) to (0.center);
		\draw [style=boldedge] (1.center) to (4);
		\draw [style=boldedge] (7.center) to (6.center);
		\draw [style=boldedge] (6.center) to (5.center);
		\draw [style=boldedge] (5.center) to (7.center);
		\draw [style=boldedge] (11.center) to (10.center);
		\draw [style=boldedge] (10.center) to (9.center);
		\draw [style=boldedge] (9.center) to (11.center);
	\end{pgfonlayer}
\end{tikzpicture} \qquad \qquad = \qquad \qquad P_{\texttt{sad}} \circ \rho_{\texttt{Alice}} \circ P_{\texttt{sad}}
\end{equation}
The grey dot involves the projection operator $P_\text{sad}$.
Every projector is a density matrix, so that the feature \texttt{sad} being imposed onto \texttt{Alice} can be illustrated as the interaction of two density matrices via the grey dot.

\citet{meaningUpdatingDensMat} introduce two mechanisms: the \textit{fuzz} \citep[originally introduced by][]{DisCoCirc, fuzz} and the \textit{phaser} \citep[originally introduced by][]{phaser1, phaser2, phaser3}. Both fuzz and phaser are guitar pedals, which are also defined by \citet{meaningUpdatingDensMat} as the mathematical composition symbols. 
The {fuzz} $\raisebox{-1.5mm}{\epsfig{figure=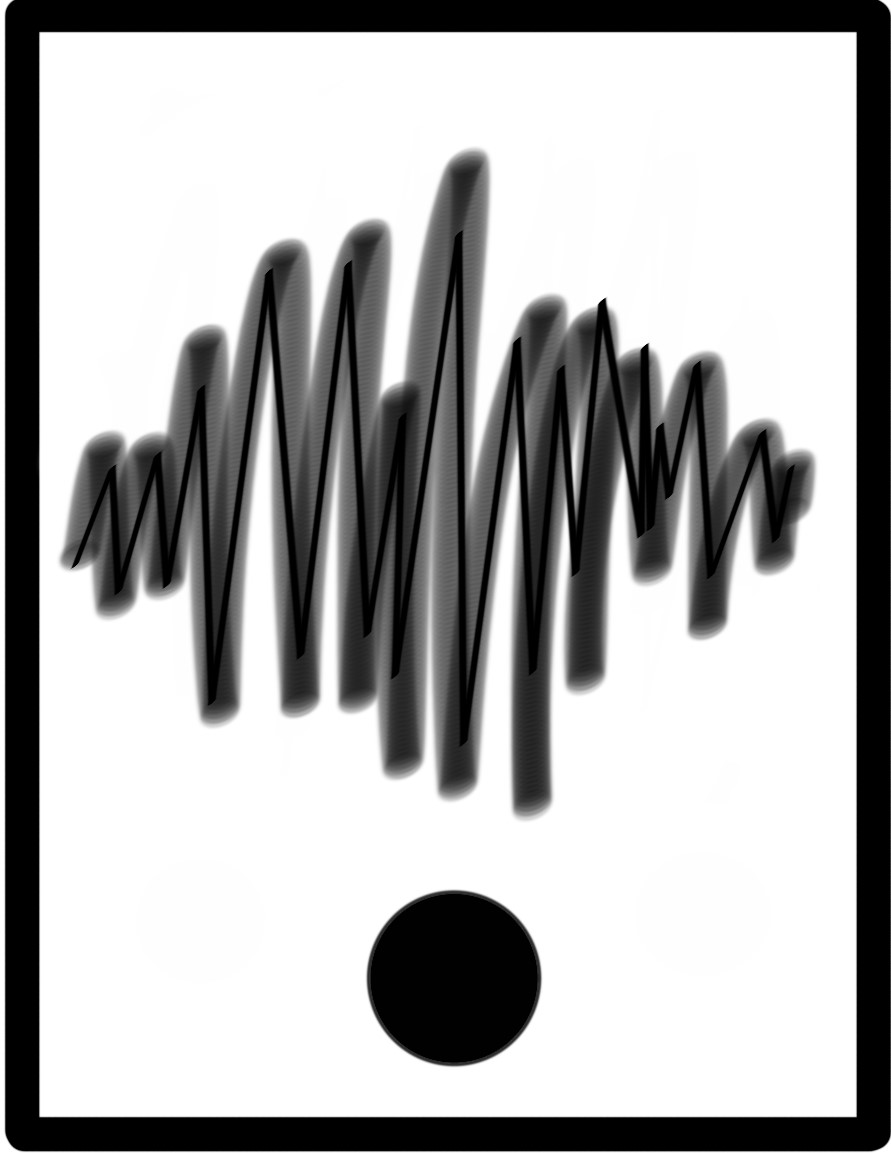,width=11pt}}$ is defined as the following operation between density matrices $\sigma$ and $\rho$:
\begin{equation}
\label{eq:fuzz}
    \rho \, \raisebox{-1.5mm}{\epsfig{figure=FUZZZZZ.jpg,width=11pt}} \, \sigma = \sum_i x_i \, \Big(P_i \circ \rho \circ P_i\Big) \qquad \qquad \text{with}\quad  \sigma := \sum_i x_i P_i
\end{equation}
whereas the phaser $\raisebox{-1.5mm}{\epsfig{figure=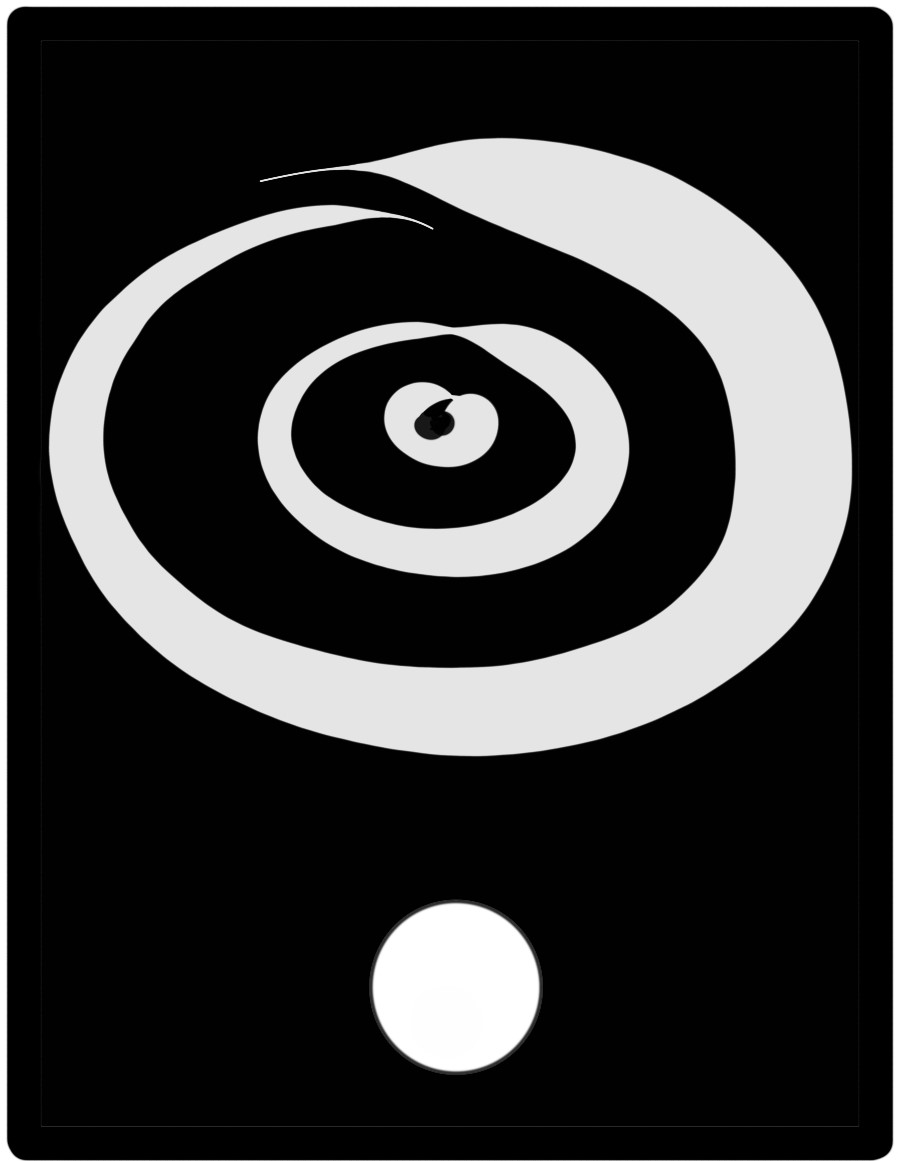,width=11pt}}$ is: 
\begin{equation}
\label{eq:phaser}
    \rho \, \raisebox{-1.5mm}{\epsfig{figure=PHASERRR.jpg,width=11pt}} \, \sigma = \Big( \sum_i x_i P_i \Big) \circ \rho \circ \Big( \sum_j x_j P_j \Big) \qquad \qquad \text{with} \quad \sigma := \sum_i x_i^2 P_i
\end{equation}
The application of either the fuzz or the phaser to update the meaning of density matrices is not a trace preserving operation. 
The normalisation of density matrices thus has to be ensured manually. 
%



In the further course of this work, we argue how mixed quantum processes can model probability distributions over \emph{functions}, starting from quantum computing and the DisCoCat framework and moving on to the DisCoCirc framework.

\section{Managing Ambiguity in Natural Language Syntax}
\label{sec:ambiguitySyntaxPaper}

\subsection{Problem Statement}


Consider again the pair of example sentences:
\begin{equation}
\label{eq:MainSentence}
    \texttt{The dog broke the vase. It was clumsy.}
\end{equation}
As mentioned in Section \ref{sec:introductionPaper}, there are three readings to this sentence. Consider the two readings of either the \texttt{dog} or the \texttt{vase} being \texttt{clumsy} for simplicity reasons. 
The \texttt{dog} being clumsy is the more likely reading: one can imagine assigning probabilities to the individual readings, subsequently creating a probability distribution over the two possible realisations. 

We model this information processing using an extension to the existing DisCoCat framework and its relation to probability distributions over quantum processes. 

\begin{figure}[!b]
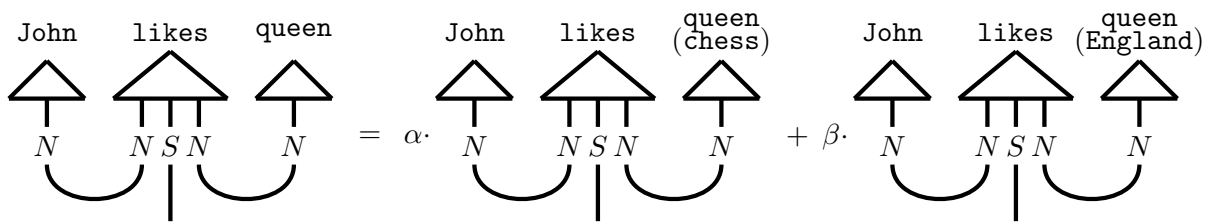

    \centering
    \ctikzfig{johnLikesQueen}
    \caption{A probability distribution over sentences, in which the noun \texttt{queen} is ambiguous, using the example sentence \texttt{John likes the queen.}}
    \label{fig:johnLikesQueen}
\end{figure}

\begin{figure}[!b]
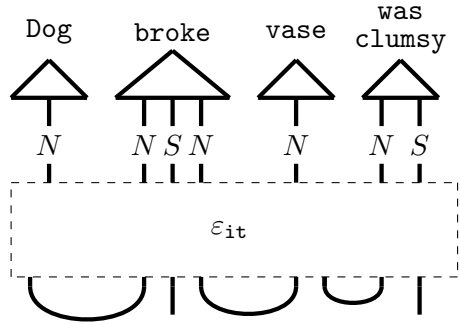

    \centering
    \ctikzfig{dogBrokeVaseMorphism}
    \caption{A diagram capturing the probability distribution over syntactically different sentences, using the example sentence \texttt{The dog broke the vase.} \texttt{It was clumsy.} $\varepsilon_{\texttt{it}}$ in the dashed box represents a probability distribution over how the words in the sentence are connected}
    \label{fig:dogBrokeVaseMorphism}
\end{figure}

Consider the sentence: 
\begin{equation}
    \texttt{John likes the queen.}
\end{equation}
Density matrices can be used to model the word \texttt{queen} as a probability distribution over \texttt{queen} \texttt{of} \texttt{England} and the chess piece \texttt{queen}, which, when plugged in, results in a probability distribution over two sentences. 
Note however the difference to Example \ref{eq:MainSentence}, in which two different readings correspond to sentences with different syntactic connections. 
In short, we extend the degree to which ambiguity in language can be modelled and move from probability distributions over sentences containing only ambiguous words (Figure \ref{fig:johnLikesQueen})
to probability distributions over different syntactic sentences (Figure \ref{fig:dogBrokeVaseMorphism}).   







\subsection{The Idea}
\label{sec:idea}

\begin{figure}
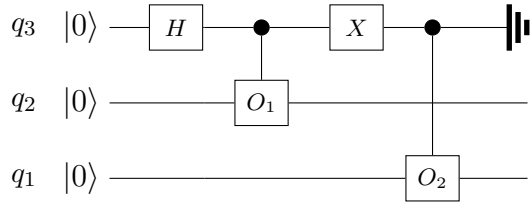

    \centering
    \ctikzfig{simpleExample}
    \caption{Ansatz for creating an equally distributed probability distribution over two operations $O_1$, $O_2$. Three qubits are used to make the calculation (Equation \ref{eq:calculationCircuit}) clearer.}
    \label{diag:mappingToQC}
\end{figure}

The goal is to obtain a diagram that models a probability distribution over different diagrams, in which the \emph{wires} connect the \emph{boxes} differently. 
We present a quantum circuit that models a probability distribution over different operations, which will later be identified with the \emph{functions} that certain words have in sentences. 
The quantum circuit, shown in Figure \ref{diag:mappingToQC} presents a probability distribution over the two operations $O_1$ and $O_2$ being executed. 
The idea behind this circuit is that, due to the $X$-gate in between the two controlled gates on qubit $q_2$, only one of the operations $O_1$ or $O_2$ is executed, both with a probability of \num{50}\%. 
The probability distribution can be adjusted from equally distributed to differently distributed, by replacing the {Hadamard} gate accordingly. 

The above statements can be verified using the explicit matrix representations for the qubits, which results in the state $\psi_{\texttt{final}}$: 
\begin{equation}
\label{eq:calculationCircuit}
\begin{split}
    \psi_{\texttt{final}} = \, &\big( \ket{0} \bra{0} \otimes 1 \otimes 1 + \ket{1}\bra{1} \otimes 1 \otimes O_2 \big) \qquad \,\,\, \textit{(controlled operation 2)}
    \\
    \circ \, &\big( \texttt{X} \otimes 1 \otimes 1 \big) \qquad \qquad \qquad \qquad \qquad \qquad \, \, \textit{(}\texttt{NOT}\textit{ gate)}
    \\
    \circ \, &\big( \ket{0} \bra{0} \otimes 1 \otimes 1 + \ket{1}\bra{1} \otimes O_1 \otimes 1 \big) \qquad \,\,\, \textit{(controlled operation 1)}
    \\
    \circ \, &\big( \texttt{H} \otimes 1 \otimes 1 \big) \qquad \qquad \qquad \qquad \qquad \qquad \,\,\, \textit{(Hadamard gate)}
    \\
    \circ \, &\big(\ket{0} \otimes \ket{0} \otimes \ket{0} \big) \qquad \qquad \qquad \qquad \qquad \textit{(initial state)}
    \\ 
\end{split}
\end{equation}
where $1$, here, is the two-dimensional identity matrix.
This expression represents the mathematical operations corresponding to carrying out the gate operations in the circuit on the qubits (Figure \ref{diag:mappingToQC}). 
Firstly, from the initial state $\ket{000}$, applying the \texttt{Hadamard} gate yields: 
\begin{equation}
    \frac{1}{\sqrt{2}} \big( \ket{0} + \ket{1} \big) \otimes \ket{0} \otimes \ket{0} 
\end{equation}
Now, applying the first controlled operation, results in: 
\begin{equation}
\begin{split}
    \frac{1}{\sqrt{2}} \big( \ket{0} \bra{0} \otimes 1 \otimes 1 + \ket{1}\bra{1} \otimes O_1 \otimes 1 \big) &\circ \big( \ket{0} \otimes \ket{0} \otimes \ket{0} + \ket{1} \otimes \ket{0} \otimes \ket{0} \big) \\= \frac{1}{\sqrt{2}} \big( \ket{0} \otimes \ket{0} \otimes \ket{0} &+ \ket{1} \otimes O_1 \ket{0} \otimes \ket{0} \big) 
\end{split}
\end{equation}
After that, the $X$-gate is applied to the first qubit, which flips the state for the first qubit. Next, the second control operation is applied:
\begin{equation}
    \frac{1}{\sqrt{2}} \big( \ket{0} \bra{0} \otimes 1 \otimes 1 + \ket{1}\bra{1} \otimes 1 \otimes O_2 \big) \circ \big( \ket{1} \otimes \ket{0} \otimes \ket{0} + \ket{0} \otimes O_1 \ket{0} \otimes \ket{0} \big)
\end{equation}
which results in the state: 
\begin{equation}
    \frac{1}{\sqrt{2}} \big( \ket{1} \otimes \ket{0} \otimes O_2 \ket{0} + \ket{0} \otimes O_1 \ket{0} \otimes \ket{0} \big)
\end{equation}
Note that, if the first operation $O_1$ is applied, the first and third qubit will be measured to $\ket{0}$, whereas if the second operation $O_2$ is applied, the first qubit will be measured to $\ket{1}$ and the second qubit will be measured to $\ket{0}$. If one operation is applied, the other is not. 
In the last step, qubit $q_2$ is discarded and a maximally mixed state is obtained, which represents a probability distribution over which of the operations are executed. 

The operations $O_1$ and $O_2$ can be associated with words in sentences that are responsible for connecting diagrams in a certain way, effectively resulting in different syntactic configurations.  
Generally, this circuit can express probability distributions that maintain linguistic ambiguity, independent of the task at hand. 

\begin{figure}
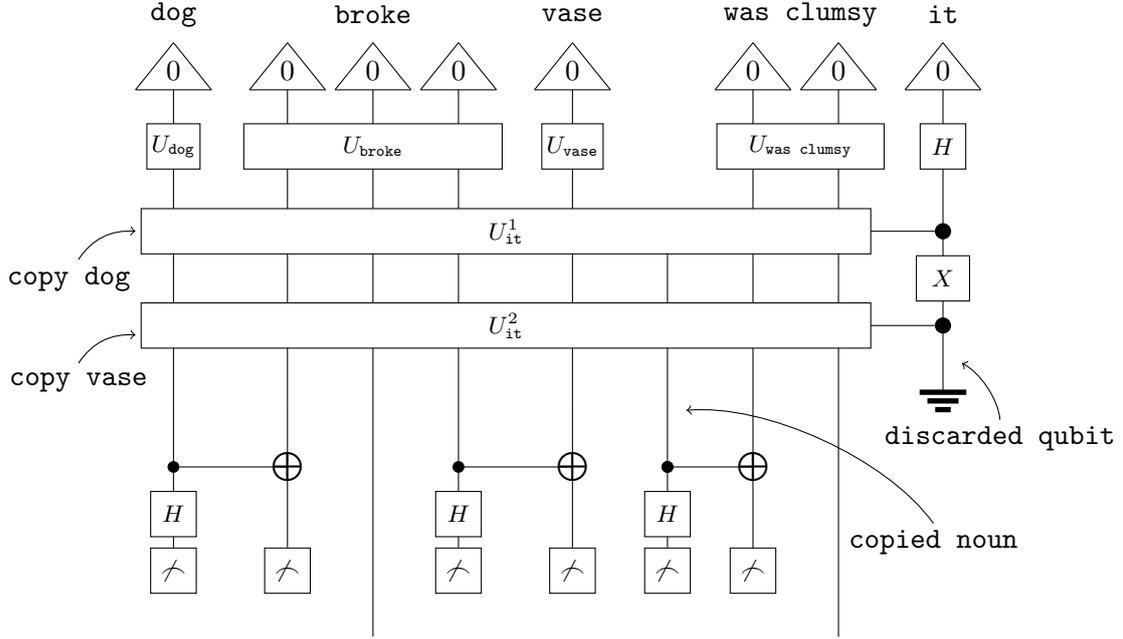

    \centering
    \ctikzfig{dogBrokeVaseQC}
    \caption{The Quantum Circuit capturing the meaning of the sentence \texttt{The dog broke the vase. It was clumsy.} The two operations captured by the unitary matrices $U_{\texttt{it}}^1$ and $U_{\texttt{it}}^2$ correspond to processes connecting the words in the circuit differently.}
    \label{fig:dogBrokeVaseQC}
\end{figure}

Consider again Example \ref{eq:MainSentence}, where the linguistic task consists of categorising the word \texttt{it} as a subject-relative pronoun or an object-relative pronoun. 
On a quantum circuit, we create the probability distribution over which one of the two is \texttt{clumsy} (Figure \ref{fig:dogBrokeVaseQC}),
where the two different operations now correspond to copying either the information of the \texttt{vase}, or the \texttt{dog}, making either one clumsy. 
Note that, in the operations represented by the unitaries $U_{\texttt{it}}^1$ and $U_{\texttt{it}}^2$, one qubit will be initialised, corresponding to the quantum circuit equivalent of the Frobenius copying map (Figure \ref{fig:fullQCdogBrokeVae}). 
On this quantum circuit, the qubit next to the qubit encoding the meaning of the word \texttt{vase} on the right, encodes the meaning of the copied noun. This copied noun will later be identified as the result of applying the Frobenius copying map, with the corresponding translation to quantum circuits. 
Here the copied noun qubit belongs to the word \texttt{it}.

We use the \texttt{Hadamard} gate to create a superposition over the states $\ket{0}$ and $\ket{1}$, resulting in an equally distributed probability distribution. 
One might want to encode previous knowledge regarding how frequently respective words appear in language, on the quantum circuit. This is possible by replacing the \texttt{Hadamard} gate with different gates, creating a different probability distribution.  

\begin{figure}
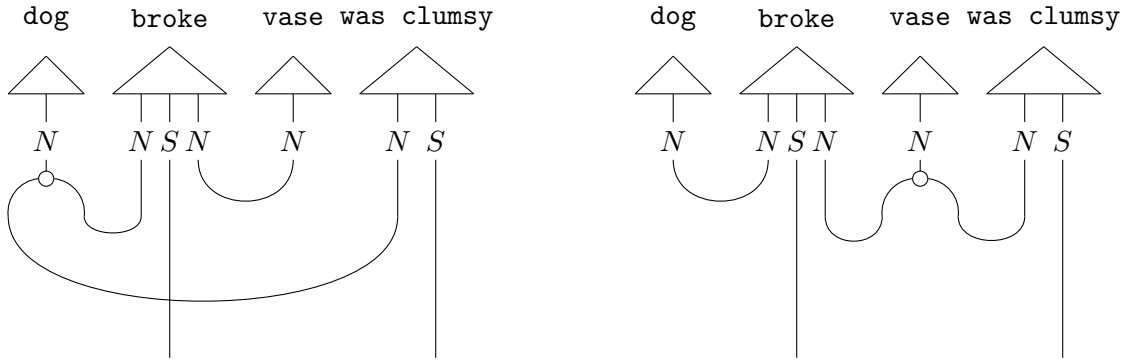

    \centering
    \ctikzfig{dogBrokeVaseOriginals}
    \caption{The diagrams capturing the respective meaning of either the \texttt{dog} (left), or the \texttt{vase} (right) being \texttt{clumsy}}
    \label{fig:dogBrokeVaseOriginals}
\end{figure}

The two different quantum circuits (Figure \ref{fig:dogBrokeVaseQC}), arising by applying either operation $U_1$ or operation $U_2$, can be associated with DisCoCat diagrams (Figure \ref{fig:dogBrokeVaseOriginals}). 
The word \texttt{it} is modelled as a probability distribution over processes. Thus, the word \texttt{it} is re-introduced as a word that carries no meaning of its own but is a probability distribution over only wires, redirecting the flow of information between the remaining words in the sentence. 
This results in the following diagrammatic equation: 
\begin{equation}
    \resizebox{0.9\textwidth}{!}{\input{dogBrokeVaseProbDist.tikz}}
\end{equation}
$\varepsilon_{\texttt{it}}$ is a probability distribution over completely positive maps: 
\begin{equation}
    \varepsilon_{\texttt{it}} = \alpha \cdot \big( f^\dag \circ - \circ f \big) + \beta \cdot \big( g^\dag \circ - \circ g \big)
\end{equation}
where:
\begin{equation}
\begin{split}
    f = 
    \, 
    &\Bigg[ 
    \underbrace{\Big( \Delta_N \otimes 1_N \otimes 1_S \otimes 1_N \otimes 1_N \otimes 1_N \otimes 1_S \Big)}_{\text{copy dog}} \\
    &\circ 
    \underbrace{\Big( 1_N \otimes \sigma_{NN} \otimes 1_S \otimes 1_N \otimes 1_N \otimes 1_N \otimes 1_S \Big)}_{\text{swap two noun-wires}} \\
    &\circ 
    \underbrace{\Big( \ldots \Big)}_{\text{\qquad \qquad \qquad swap remaining and compose\qquad \qquad \qquad}} \Bigg]\\
\end{split}
\end{equation}
where $\sigma_{NN}$ is a morphism that swaps the noun-wires \citep{coeckebook}. The morphism $g$ corresponds to copying the word \texttt{vase} and connecting the words in the sentence accordingly. 
The other words are {pure} quantum states that the word \texttt{it} connects to in different ways. 
We can, using the CNOT gate (the quantum way of modelling Frobenius copying maps, Equation \ref{eq:spiderCnot}) obtain the full quantum circuit (Figure \ref{fig:fullQCdogBrokeVae}). 
\begin{figure}[!b]
    \centering
    \ctikzfig{fullQCdogBrokeVase}
    \caption{The full resulting quantum circuit for the sentences \texttt{The dog broke the vase. It was clumsy.}}
    \label{fig:fullQCdogBrokeVae}
\end{figure}

\subsection{The Reasoning Process}
\label{sec:ReasoningProcessPaper}

The aforementioned method works for all sentence pairs that follow the structure: 
\[\texttt{Subject verb object. It verb* adjective.}\]
The process introduced above can thus be used to train a model to differentiate between subject- and object-relative pronouns in sentences. \citet{pronounResolution, qnlpInPractice} explicitly trained VQCs for the task of differentiating between subject- and object-relative pronoun sentences. 
If a model is trained on a semantic task, such as teaching the model how closely words are related to each other, the model is implicitly trained to differentiate between these two different types of sentences. 

Note that the model always learns the meaning of sentences in terms of the sentence space. In many cases, the sentence space is subject to a classification task. This could be a binary classification task, and thus a two-dimensional sentence space containing vectors representing the meanings of \texttt{True} and \texttt{False}, for example. 
\citet{qnlpInPractice} define these categories as \texttt{subject relative pronoun sentence} and \texttt{object relative pronoun sentence} and the categories \texttt{food} and \texttt{IT}, in a second task.
The sentence space is always restricted to the categories that the model has been trained on. 




Sticking to the two-dimensional case, one approach might be choosing the categories \texttt{True} and \texttt{False} and, in connection with Example \ref{eq:MainSentence}, teaching the model that a \texttt{vase} cannot be \texttt{clumsy} (\texttt{vase is clumsy} $\rightarrow$ \texttt{False}), while a \texttt{dog} can (\texttt{dog is clumsy} $\rightarrow$ \texttt{True}).
Then, when a probability distribution is created over the two possibilities of the \texttt{dog} or the \texttt{vase} being clumsy, ultimately, a probability distribution is created over the model predicting \texttt{True} or \texttt{False}.

The reasoning process entails using \emph{projection operators} to project out the \texttt{True} part of the prediction. 
The upside of this method is that, on a small scale, the \textit{syntactic} connections can be recovered from the \emph{semantic} connections that the model learned. 
Consider the spectrally decomposed form of a density matrix: 
\begin{equation}
    \sigma = \sum_i x_i P_i
\end{equation}
where $P_i$ are projection operators, weighted by the terms $x_i$. 
Consider now two sentences, $\rho_{\texttt{s1}} = 
\sum_i y_i P_i$, and $\rho_{\texttt{s2}} = \sum_i x_i P_i$, which might both be ambiguous, so that, using the {fuzz} (Equation \ref{eq:fuzz}):
\begin{equation}
\begin{split}
    \rho_{s1} \raisebox{-1.5mm}{\epsfig{figure=FUZZZZZ.jpg,width=11pt}} \rho_{s2} &= \sum_i \sum_j x_i y_j \Big( \ket{i} \braket{i|j} \braket{j|i} \bra{i}\Big) \\
    &= \sum_i \sum_j x_i y_j \cdot |\delta_{i, j}|^2 \ket{i} \bra{i} \\
    &= \sum_i x_i y_i \ket{i} \bra{i} 
\end{split}
\end{equation}
where $\delta_{i, j}$ is the Kronecker-Delta. The result is another (not necessarily normalised) density matrix, representing the updated meaning of the sentence.
In the context of Example \ref{eq:MainSentence}, the goal is to teach the model that the words \texttt{dog} and \texttt{clumsy} are closer together than the words \texttt{vase} and \texttt{clumsy}. 

To achieve this, one assigns an appropriate meaning to the sentence space, such as \texttt{True} and \texttt{False} and the representation of the sentence is a probability distribution over these two categories. This allows for the use of projectors (Section \ref{sec:DisCoCatPaper}) to recover (with a certain probability) the diagram yielding the meaning \texttt{True} or \texttt{False}. 

\subsection{Verb Phrase Ellipsis}

\begin{figure}
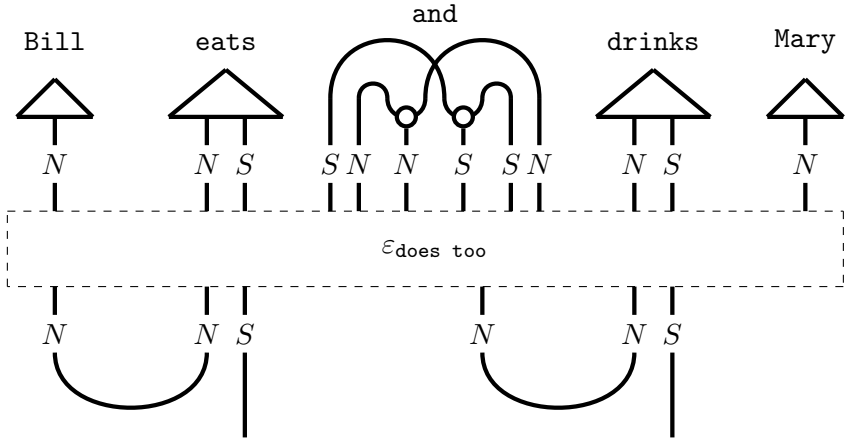

    \centering
    \ctikzfig{billEatsDrinksMaryDoesTooMorphism}
    \caption{The diagram modelling a probability distribution over the different sentence meanings, for the sentences \texttt{Bill eats and drinks. Mary does too.} The possible internal wirings for the phrase \texttt{does too} is displayed in Figure \ref{fig:doesTooInternalWiring}.}
    \label{fig:billEatsDrinksMaryDoesTooMorphism}
\end{figure}

\begin{figure}
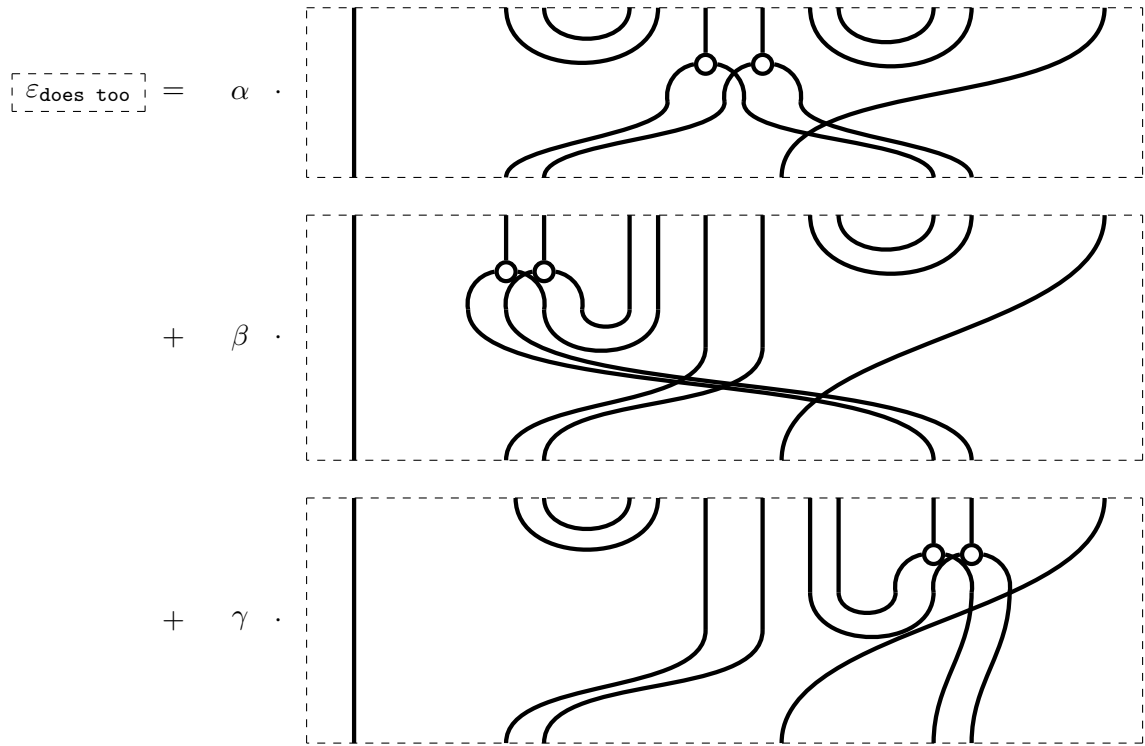

    \centering
    \ctikzfig{doesTooInternalWiring}
    \caption{The internal wirings for the phrase \texttt{does} \texttt{too} in Figure \ref{fig:billEatsDrinksMaryDoesTooMorphism}. The first diagram represents \texttt{\ldots Mary eats and drinks}, the second one represents \texttt{\ldots Mary eats} and the third one represents \texttt{\ldots Mary drinks}}
    \label{fig:doesTooInternalWiring}
\end{figure}

Consider the case of {verb phrase ellipsis}, as treated by \citet{Wijnholds_2020}. 
\emph{Ellipsis} generally describes one or more words missing in a sentence \citep{ellipsis}. In the case of verb phrase ellipsis, the missing words constitute a verb phrase. 
In his work, \citet{Wijnholds_2020} introduces a controlled way of copying information and inserting it in the appropriate place in the sentence, using an extension to the {Lambek calculus}.
Consider the example sentence: 
\begin{displayquote}
    \texttt{Bill eats and drinks. Mary does too.}
\end{displayquote}
with the three possible readings:
\begin{enumerate}\itemsep=0pt
    \item \texttt{Bill eats and drinks. Mary eats.}
    \item \texttt{Bill eats and drinks. Mary drinks.}
    \item \texttt{Bill eats and drinks. Mary eats and drinks.}
\end{enumerate}
\citet{Wijnholds_2020} models the phrase \texttt{does} \texttt{too} as: 
\begin{equation}
    \begin{tikzpicture}
	\begin{pgfonlayer}{nodelayer}
		\node [style=none] (0) at (-0.75, 0.5) {};
		\node [style=none] (1) at (0.75, 0.5) {};
		\node [style=none] (2) at (-1.5, 0.5) {};
		\node [style=none] (3) at (1.5, 0.5) {};
		\node [style=none] (4) at (0, 2.5) {\texttt{does too}};
		\node [style=none] (5) at (-1.5, -0.25) {};
		\node [style=none] (6) at (-0.75, -0.25) {};
		\node [style=none] (7) at (0.75, -0.25) {};
		\node [style=none] (8) at (1.5, -0.25) {};
		\node [style=none] (9) at (1.5, -0.75) {$S$};
		\node [style=none] (10) at (-1.5, -0.75) {$S$};
		\node [style=none] (11) at (-1.5, -1.25) {};
		\node [style=none] (12) at (1.5, -1.25) {};
		\node [style=none] (13) at (-0.75, -0.75) {$N$};
		\node [style=none] (14) at (0.75, -0.75) {$N$};
		\node [style=none] (15) at (-0.75, -1.25) {};
		\node [style=none] (16) at (0.75, -1.25) {};
	\end{pgfonlayer}
	\begin{pgfonlayer}{edgelayer}
		\draw [style=none, bend left=90, looseness=1.75] (0.center) to (1.center);
		\draw [style=none, bend left=90, looseness=1.75] (2.center) to (3.center);
		\draw [style=none] (6.center) to (0.center);
		\draw [style=none] (2.center) to (5.center);
		\draw [style=none] (7.center) to (1.center);
		\draw [style=none] (3.center) to (8.center);
	\end{pgfonlayer}
\end{tikzpicture}    
\end{equation}
together with the appropriate verb being copied. 
Now, the probability distribution can be constructed, yielding the three original readings (Figure \ref{fig:billEatsDrinksMaryDoesTooMorphism}). 
Again, reducing the different ways of connecting the words into a probability distribution over morphisms, captured by $\varepsilon_{\texttt{does too}}$. 
$\varepsilon_{\texttt{does too}}$ is a probability distribution over the individual tensor products of processes (Figure \ref{fig:doesTooInternalWiring}).
We thus expand the theory by \citet{Wijnholds_2020} and capture all the nonlinearity in the phrase \texttt{does too}.

\begin{figure}
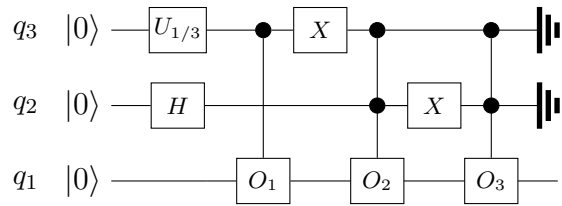

    \centering
    \ctikzfig{threeOperationsCircuit}
    \caption{The quantum circuit to create an equally distributed probability distribution over three operations on qubit $q_1$. $U_{\sqrt{1/3}}$ is a unitary matrix creating a probability distribution as: $\sqrt{1/3} \ket{1} + \sqrt{2/3} \ket{0}$}
    \label{fig:probDistOverThreeQC}
\end{figure}

\begin{figure}
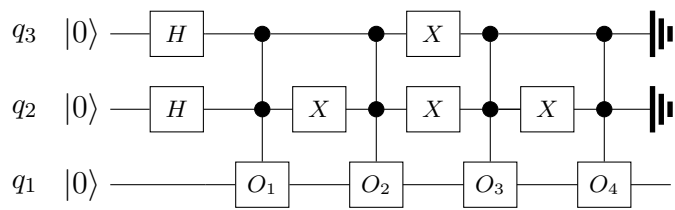

    \centering
    \ctikzfig{simpleExampleFour}
    \caption{The quantum circuit to create an equally distributed probability distribution over four operations on qubit $q_1$}
    \label{fig:probDistOverFourQC}
\end{figure}

The phrase \texttt{does} \texttt{too} creates ambiguity in the sentence and can be displayed entirely in terms of wires, and can thus be understood as the \emph{source} of ambiguity in the sentence. 
The arising quantum circuit is different in the sense that here, we have a probability distribution over three operations, which can be accounted for by entangling individual qubits with each other, which are then discarded (Figure \ref{fig:probDistOverThreeQC}). One can achieve a probability distribution over four operations similarly (Figure \ref{fig:probDistOverFourQC}).

\begin{figure}
    \centering
    \resizebox{0.7\textwidth}{!}{\input{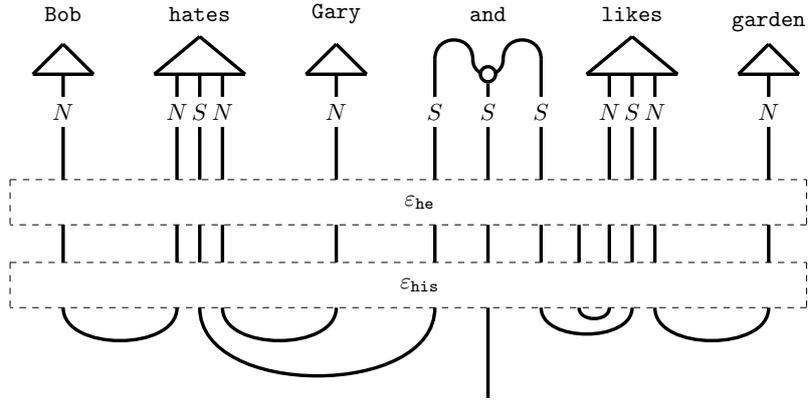}}
    \caption{The diagram representing the probability distribution, caused by the sentence \texttt{Bob hates Gary, and he likes his garden.} There are four possible readings to this sentence, all captured by this diagram.}
    \label{fig:bobLikesGaryAndSoOn}
\end{figure}

The sentence:
\begin{equation}
\label{eq:BobGaryGardenSentence}
    \texttt{Bob hates Gary and he likes his garden.}
\end{equation}
contains ambiguity as to who likes whose \texttt{garden}. 
Our model captures this ambiguity as well, as long as the two individual sentences (in particular the words \texttt{he} and \texttt{his}) can be modelled as DisCoCat diagrams.
The meaning of this sentence can be captured by a diagram, where the explicit inner wirings for the words \texttt{he} and \texttt{his} are omitted (Figure \ref{fig:bobLikesGaryAndSoOn}). 
Note that in Example \ref{eq:BobGaryGardenSentence}, replacing the word \texttt{and} with the word \texttt{but} would eliminate the ambiguity ({it} would be \texttt{Bob} that likes \texttt{Gary}'s garden).

\subsection{Ambiguity in Interaction}

\begin{figure}
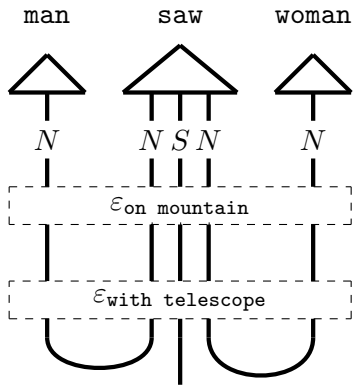

    \centering
    \ctikzfig{manWomanTelescope}
    \caption{The diagram encoding a probability distribution, where there are two sources of ambiguity, caused by the phrases \texttt{on mountain} and \texttt{with telescope}. The arising probability distribution consists of four individual terms.}
    \label{fig:manWomanTelescope}
\end{figure}

Consider introducing two sources of ambiguity: 
\begin{equation}
    \texttt{The man saw the woman on the mountain with a telescope.}
\end{equation}
There are at least nine readings for this sentence: 
\begin{enumerate}\itemsep=0pt
    \item {The (man on the mountain) saw the (woman with a telescope).}
    \item {The (man with a telescope) saw the (woman on the mountain).}
    \item {The (man on the mountain with a telescope) saw the woman.}
    \item {The man saw the (woman on the mountain with a telescope).}
    \item {The (man on the mountain) (saw with a telescope) the woman.}
    \item {The man (saw with a telescope) the (woman on the mountain).}
    \item {The man and the woman are on the mountain. The man saw the (woman with a telescope).}
    \item {The man and the woman are on the mountain. The (man with a telescope) saw the woman.}
    \item {The man and the woman are on the mountain. The man (saw with a telescope) the woman.}
\end{enumerate}
Note that \texttt{with} \texttt{a} \texttt{telescope} can either mean \texttt{who} \texttt{has} \texttt{a} \texttt{telescope} or \texttt{use} \texttt{a} \texttt{telescope} \texttt{to} \texttt{do} \texttt{something}. 
Following the reasoning above, all of these nine readings can be modelled within one diagram (Figure \ref{fig:manWomanTelescope}), where, in one of the cases, the phrase \texttt{with telescope} acts as a verb modifier on the verb \texttt{saw}, and the nouns are simple wires.
The explicit inner wirings for the two probability distributions are according to the following diagrammatic equation: 
\begin{equation}
\label{eq:innerWiringManWomanTelescope}
    \begin{tikzpicture}
	\begin{pgfonlayer}{nodelayer}
		\node [style=none] (92) at (-8, 0) {$\varepsilon_{\texttt{on mountain}}$};
		\node [style=none] (93) at (-5.5, 0) {$=$};
		\node [style=none] (94) at (-4.25, 0) {$\alpha$};
		\node [style=none] (95) at (-3.75, 0) {$\cdot$};
		\node [style=none] (96) at (-9.75, 0.5) {};
		\node [style=none] (97) at (-9.75, -0.5) {};
		\node [style=none] (98) at (-6.25, 0.5) {};
		\node [style=none] (99) at (-6.25, -0.5) {};
		\node [style=none] (100) at (1.75, 3) {};
		\node [style=none] (101) at (1, 3) {};
		\node [style=none] (102) at (2.5, 3) {};
		\node [style=none] (103) at (1.75, -1.75) {};
		\node [style=none] (104) at (1.75, -1.75) {};
		\node [style=none] (105) at (1.75, -1.75) {};
		\node [style=none] (106) at (1.75, -1.75) {};
		\node [style=none] (107) at (1, -1.75) {};
		\node [style=none] (108) at (1, -1.75) {};
		\node [style=none] (109) at (1, -1.75) {};
		\node [style=none] (110) at (1, -1.75) {};
		\node [style=none] (111) at (2.5, -1.75) {};
		\node [style=none] (112) at (2.5, -1.75) {};
		\node [style=none] (113) at (2.5, -1.75) {};
		\node [style=none] (114) at (2.5, -1.75) {};
		\node [style=none] (115) at (-1.5, 1.25) {};
		\node [style=none] (116) at (-0.5, 0.5) {};
		\node [style=none] (117) at (-2.5, 0.5) {};
		\node [style=none] (118) at (-2, 0.5) {};
		\node [style=none] (119) at (-2, -0.5) {$N$};
		\node [style=none] (120) at (-2, 0) {};
		\node [style=none] (121) at (-1, 0.5) {};
		\node [style=none] (122) at (-1, -0.5) {$N$};
		\node [style=none] (123) at (-1, 0) {};
		\node [style=none] (124) at (-1.5, 2.25) {\texttt{on}};
		\node [style=none] (125) at (-1.5, 1.75) {\texttt{mountain}};
		\node [style=none] (126) at (0.25, 3) {};
		\node [style=none] (127) at (0.25, -0.5) {};
		\node [style=none] (128) at (0.25, -0.5) {};
		\node [style=none] (129) at (0.25, -0.5) {};
		\node [style=none] (130) at (0.25, -1) {};
		\node [style=none] (131) at (-1, -1) {};
		\node [style=none] (132) at (-2, -1) {};
		\node [style=none] (133) at (-2, -1.75) {};
		\node [style=none] (134) at (3.25, 3) {};
		\node [style=none] (135) at (3.25, -1.75) {};
		\node [style=none] (136) at (3.25, -1.75) {};
		\node [style=none] (137) at (3.25, -1.75) {};
		\node [style=none] (138) at (3.25, -1.75) {};
		\node [style=none] (139) at (12, 1.25) {};
		\node [style=none] (140) at (13, 0.5) {};
		\node [style=none] (141) at (11, 0.5) {};
		\node [style=none] (142) at (11.5, 0.5) {};
		\node [style=none] (143) at (11.5, -0.5) {$N$};
		\node [style=none] (144) at (11.5, 0) {};
		\node [style=none] (145) at (12.5, 0.5) {};
		\node [style=none] (146) at (12.5, -0.5) {$N$};
		\node [style=none] (147) at (12.5, 0) {};
		\node [style=none] (148) at (12, 2.25) {\texttt{on}};
		\node [style=none] (149) at (12, 1.75) {\texttt{mountain}};
		\node [style=none] (150) at (8.75, 3) {};
		\node [style=none] (151) at (8, 3) {};
		\node [style=none] (152) at (9.5, 3) {};
		\node [style=none] (153) at (8.75, -1.75) {};
		\node [style=none] (154) at (8.75, -1.75) {};
		\node [style=none] (155) at (8.75, -1.75) {};
		\node [style=none] (156) at (8.75, -1.75) {};
		\node [style=none] (157) at (8, -1.75) {};
		\node [style=none] (158) at (8, -1.75) {};
		\node [style=none] (159) at (8, -1.75) {};
		\node [style=none] (160) at (8, -1.75) {};
		\node [style=none] (161) at (9.5, -1.75) {};
		\node [style=none] (162) at (9.5, -1.75) {};
		\node [style=none] (163) at (9.5, -1.75) {};
		\node [style=none] (164) at (9.5, -1.75) {};
		\node [style=none] (165) at (7.25, 3) {};
		\node [style=none] (166) at (7.25, -1.75) {};
		\node [style=none] (167) at (7.25, -1.75) {};
		\node [style=none] (168) at (7.25, -1.75) {};
		\node [style=none] (169) at (10.25, 3) {};
		\node [style=none] (170) at (10.25, -0.5) {};
		\node [style=none] (171) at (10.25, -0.5) {};
		\node [style=none] (172) at (10.25, -0.5) {};
		\node [style=none] (173) at (10.25, -1) {};
		\node [style=none] (174) at (11.5, -1) {};
		\node [style=none] (175) at (10.25, -1) {};
		\node [style=none] (176) at (12.5, -1) {};
		\node [style=none] (177) at (12.5, -1.75) {};
		\node [style=none] (178) at (5.5, 0) {$\beta$};
		\node [style=none] (179) at (6, 0) {$\cdot$};
		\node [style=none] (180) at (-3.25, 3) {};
		\node [style=none] (181) at (-3.25, -1.75) {};
		\node [style=none] (182) at (4, 3) {};
		\node [style=none] (183) at (4, -1.75) {};
		\node [style=none] (184) at (6.5, 3) {};
		\node [style=none] (185) at (6.5, -1.75) {};
		\node [style=none] (186) at (13.75, 3) {};
		\node [style=none] (187) at (13.75, -1.75) {};
		\node [style=none] (188) at (4.5, 0) {$+$};
	\end{pgfonlayer}
	\begin{pgfonlayer}{edgelayer}
		\draw [style=dashed] (96.center) to (98.center);
		\draw [style=dashed] (98.center) to (99.center);
		\draw [style=dashed] (99.center) to (97.center);
		\draw [style=dashed] (97.center) to (96.center);
		\draw [style=doubled] (104.center) to (106.center);
		\draw [style=doubled] (108.center) to (110.center);
		\draw [style=doubled] (112.center) to (114.center);
		\draw [style=doubled] (110.center) to (101.center);
		\draw [style=doubled] (106.center) to (100.center);
		\draw [style=doubled] (114.center) to (102.center);
		\draw [style=doubled] (117.center) to (116.center);
		\draw [style=doubled] (116.center) to (115.center);
		\draw [style=doubled] (115.center) to (117.center);
		\draw [style=doubled] (118.center) to (120.center);
		\draw [style=doubled] (121.center) to (123.center);
		\draw [style=doubled] (128.center) to (130.center);
		\draw [style=doubled] (130.center) to (126.center);
		\draw [style=doubled, bend right=90] (131.center) to (130.center);
		\draw [style=doubled] (133.center) to (132.center);
		\draw [style=doubled] (136.center) to (138.center);
		\draw [style=doubled] (138.center) to (134.center);
		\draw [style=doubled] (141.center) to (140.center);
		\draw [style=doubled] (140.center) to (139.center);
		\draw [style=doubled] (139.center) to (141.center);
		\draw [style=doubled] (142.center) to (144.center);
		\draw [style=doubled] (145.center) to (147.center);
		\draw [style=doubled] (154.center) to (156.center);
		\draw [style=doubled] (158.center) to (160.center);
		\draw [style=doubled] (162.center) to (164.center);
		\draw [style=doubled] (160.center) to (151.center);
		\draw [style=doubled] (156.center) to (150.center);
		\draw [style=doubled] (164.center) to (152.center);
		\draw [style=doubled] (171.center) to (173.center);
		\draw [style=doubled] (173.center) to (169.center);
		\draw [style=doubled] (168.center) to (165.center);
		\draw [style=doubled, bend right=90] (175.center) to (174.center);
		\draw [style=doubled] (177.center) to (176.center);
		\draw [style=dashed] (180.center) to (182.center);
		\draw [style=dashed] (182.center) to (183.center);
		\draw [style=dashed] (183.center) to (181.center);
		\draw [style=dashed] (181.center) to (180.center);
		\draw [style=dashed] (184.center) to (186.center);
		\draw [style=dashed] (186.center) to (187.center);
		\draw [style=dashed] (187.center) to (185.center);
		\draw [style=dashed] (185.center) to (184.center);
	\end{pgfonlayer}
\end{tikzpicture}
\end{equation}
where we omitted the action of the words \texttt{on} \texttt{mountain} and \texttt{with} \texttt{telescope} on the verbs, for simplicity reasons and the same diagrammatic reasoning works for the phrase \texttt{with} \texttt{telescope}.
This results in a density matrix with four terms, displaying four possible readings of the ultimate sentence.

In the explicit representation (Equation \ref{eq:innerWiringManWomanTelescope}), $\ket{\texttt{on mountain}}$ is a state, but according to the \emph{process-state-duality} \citep[][]{coeckebook}\footnote{This has a direct equivalent in quantum theory: the \textit{Choi-Jamiołkowski isomorphism} \citep{choi, jamiolkowski}, which is also referred to as the \textit{channel-state duality}}, it can be viewed as a process, making $\varepsilon_{\texttt{on mountain}}$ a probability distribution over processes. 

The more involved case of the phrases \texttt{on} \texttt{mountain} and \texttt{with} \texttt{telescope} being verb modifiers are as straightforward and we will account for these in the DisCoCirc framework in the next section. 

\subsection{Extension to DisCoCirc}

\begin{figure}[h]
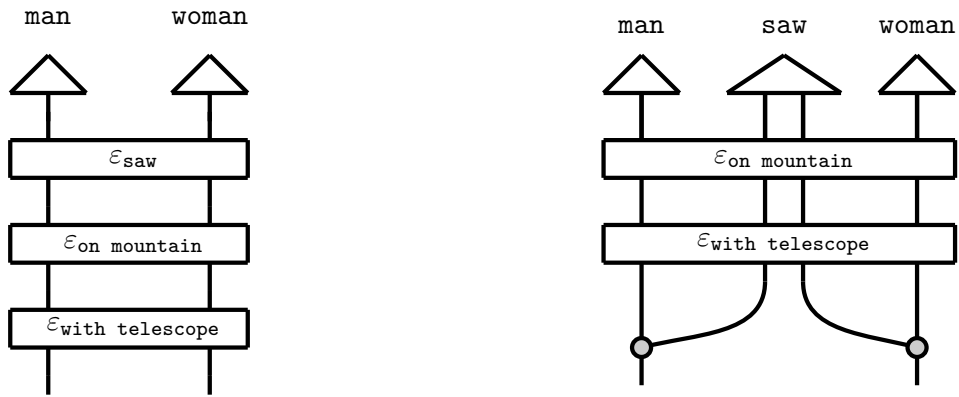

\begin{center}
\begin{minipage}{0.45\textwidth}
\centering
\ctikzfig{manWomanDisCoCirc}
\end{minipage}
\hfill
\begin{minipage}{0.45\textwidth}
\centering
\ctikzfig{dynamicVerb}
\end{minipage}
\caption{The meaning of the sentence \texttt{The man saw the woman on the mountain with a telescope} in the DisCoCirc framework (left) and the introduction of {dynamic verbs} in the DisCoCirc framework (right)}
\label{fig:manWomanDisCoCircDynamicVerbMinipage}
\end{center}
\end{figure}

\begin{figure}[h]
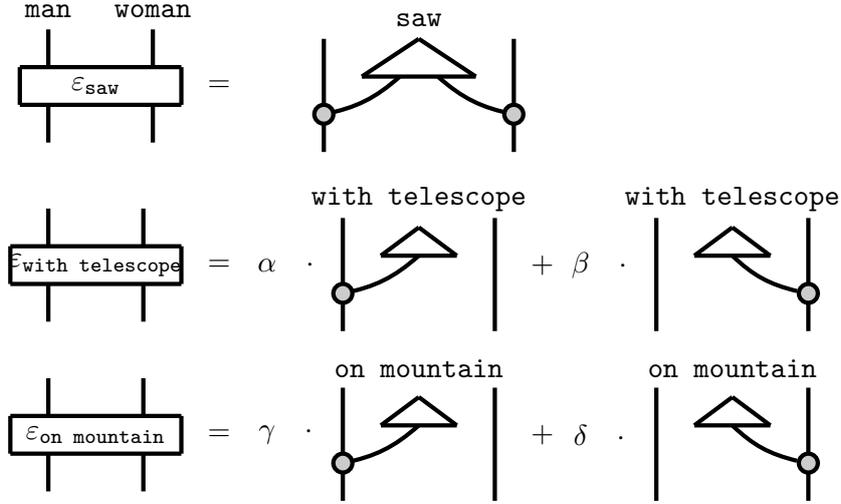

    \centering
    \ctikzfig{saw}
    \caption{The internal wirings of the individual words, represented by probability distributions over processes (Figure \ref{fig:manWomanDisCoCircDynamicVerbMinipage}, left)}
    \label{fig:saw}
\end{figure}

Using only dynamic nouns, we can model the meaning of the above sentence (Figure \ref{fig:manWomanDisCoCircDynamicVerbMinipage}, left). Individual internal wirings can be associated with these diagrams (Figure \ref{fig:saw}). 
We suggest, within the DisCoCirc framework, the extension to \emph{dynamic verbs} (Figure \ref{fig:manWomanDisCoCircDynamicVerbMinipage}, right).
Here, the scope of the dynamic verb is ended by hand, after it has been updated. One could also include this ending of the scope in the updating mechanisms themselves (Equation \ref{eq:dynamicVerbScope}).
\begin{equation}
\label{eq:dynamicVerbScope}
    \resizebox{0.9\textwidth}{!}{\begin{tikzpicture}
	\begin{pgfonlayer}{nodelayer}
		\node [style=none] (0) at (-14.5, -0.5) {};
		\node [style=none] (1) at (-19.5, -0.5) {};
		\node [style=none] (2) at (-19.5, 0.5) {};
		\node [style=none] (3) at (-14.5, 0.5) {};
		\node [style=none] (4) at (-16.75, 0) {$\varepsilon_{\texttt{with telescope}}$};
		\node [style=none] (5) at (-18.5, 0.5) {};
		\node [style=none] (6) at (-15.5, 0.5) {};
		\node [style=none] (7) at (-18.5, -0.5) {};
		\node [style=none] (8) at (-15.5, -0.5) {};
		\node [style=none] (9) at (-18.5, 1.5) {};
		\node [style=none] (10) at (-15.5, 1.5) {};
		\node [style=none] (11) at (-18.5, -1.5) {};
		\node [style=none] (12) at (-15.5, -1.5) {};
		\node [style=none] (13) at (-13.5, 0) {$=$};
		\node [style=none] (14) at (-8.25, 1.25) {};
		\node [style=none] (15) at (-7.25, 0.5) {};
		\node [style=none] (16) at (-9.25, 0.5) {};
		\node [style=none] (17) at (-8.25, 0.5) {};
		\node [style=none] (18) at (-10.25, -0.5) {};
		\node [style=none] (19) at (-10.25, -0.5) {};
		\node [style=none] (20) at (-10.25, 1.5) {};
		\node [style=none] (21) at (-10.25, -2) {};
		\node [style=none] (22) at (-5, -2) {};
		\node [style=none] (23) at (-5, 1.5) {};
		\node [style=none] (24) at (-5, -2) {};
		\node [style=none] (25) at (-7.25, 0.5) {};
		\node [style=none] (26) at (-8.25, 2) {\texttt{with telescope}};
		\node [style=gray ddot] (27) at (-10.25, -0.5) {};
		\node [style=none] (28) at (-11.75, 0) {$\alpha \quad \cdot$};
		\node [style=none] (29) at (2.5, 1.25) {};
		\node [style=none] (30) at (3.5, 0.5) {};
		\node [style=none] (31) at (1.5, 0.5) {};
		\node [style=none] (32) at (2.5, 0.5) {};
		\node [style=none] (33) at (-1, -2) {};
		\node [style=none] (34) at (-1, 1.5) {};
		\node [style=none] (35) at (-1, -2) {};
		\node [style=none] (36) at (4.5, -2) {};
		\node [style=none] (37) at (4.5, 1.5) {};
		\node [style=none] (38) at (4.5, -2) {};
		\node [style=none] (39) at (3.5, 0.5) {};
		\node [style=none] (40) at (2.5, 2) {\texttt{with telescope}};
		\node [style=gray ddot] (41) at (4.5, -0.5) {};
		\node [style=none] (42) at (-2.5, 0) {$\beta \quad \cdot$};
		\node [style=none] (43) at (-4, 0) {$+$};
		\node [style=none] (44) at (5.75, 0) {$+$};
		\node [style=none] (45) at (-17.25, 0.5) {};
		\node [style=none] (47) at (-17.25, 1.5) {};
		\node [style=none] (49) at (-16.75, 0.5) {};
		\node [style=none] (51) at (-16.75, 1.5) {};
		\node [style=none] (54) at (-6.5, -0.25) {};
		\node [style=none] (56) at (-6.5, 1.5) {};
		\node [style=none] (58) at (-6, -0.25) {};
		\node [style=none] (60) at (-6, 1.5) {};
		\node [style=none] (61) at (0, -0.25) {};
		\node [style=none] (62) at (0, 1.5) {};
		\node [style=none] (63) at (0.5, -0.25) {};
		\node [style=none] (64) at (0.5, 1.5) {};
		\node [style=none] (65) at (4.5, -1.5) {};
		\node [style=none] (66) at (0.5, -0.25) {};
		\node [style=gray ddot] (67) at (4.5, -1.5) {};
		\node [style=none] (68) at (0.5, -0.25) {};
		\node [style=none] (69) at (0, -0.25) {};
		\node [style=none] (70) at (-1, -1.5) {};
		\node [style=gray ddot] (71) at (-1, -1.5) {};
		\node [style=none] (72) at (0, -0.25) {};
		\node [style=none] (73) at (-6.5, -0.25) {};
		\node [style=none] (74) at (-10.25, -1.5) {};
		\node [style=gray ddot] (75) at (-10.25, -1.5) {};
		\node [style=none] (76) at (-6.5, -0.25) {};
		\node [style=none] (77) at (-5, -1.5) {};
		\node [style=none] (78) at (-6, -0.25) {};
		\node [style=gray ddot] (79) at (-5, -1.5) {};
		\node [style=none] (80) at (-6, -0.25) {};
		\node [style=none] (81) at (7, 0) {$\gamma \quad \cdot$};
		\node [style=none] (82) at (11, 1.5) {};
		\node [style=none] (83) at (12, 0.75) {};
		\node [style=none] (84) at (10, 0.75) {};
		\node [style=none] (86) at (8.5, -2) {};
		\node [style=none] (87) at (8.5, 1.25) {};
		\node [style=none] (88) at (8.5, -2) {};
		\node [style=none] (89) at (14, -2) {};
		\node [style=none] (90) at (14, 1.25) {};
		\node [style=none] (91) at (14, -2) {};
		\node [style=none] (92) at (12, 0.75) {};
		\node [style=none] (93) at (11, 2) {\texttt{with telescope}};
		\node [style=none] (95) at (9.5, -0.25) {};
		\node [style=none] (96) at (9.5, 1.25) {};
		\node [style=none] (97) at (12.5, -0.25) {};
		\node [style=none] (98) at (12.5, 1.25) {};
		\node [style=none] (99) at (14, -1.5) {};
		\node [style=none] (100) at (12.5, -0.25) {};
		\node [style=gray ddot] (101) at (14, -1.5) {};
		\node [style=none] (102) at (12.5, -0.25) {};
		\node [style=none] (103) at (9.5, -0.25) {};
		\node [style=none] (104) at (8.5, -1.5) {};
		\node [style=gray ddot] (105) at (8.5, -1.5) {};
		\node [style=none] (106) at (9.5, -0.25) {};
		\node [style=none] (107) at (10.5, 0.75) {};
		\node [style=none] (108) at (11.5, 0.75) {};
		\node [style=none] (109) at (12.5, -0.25) {};
		\node [style=none] (110) at (11.5, 0.75) {};
		\node [style=gray ddot] (111) at (12.5, -0.25) {};
		\node [style=none] (112) at (11.5, 0.75) {};
		\node [style=none] (113) at (10.5, 0.75) {};
		\node [style=none] (114) at (9.5, -0.25) {};
		\node [style=gray ddot] (115) at (9.5, -0.25) {};
		\node [style=none] (116) at (10.5, 0.75) {};
	\end{pgfonlayer}
	\begin{pgfonlayer}{edgelayer}
		\draw [style=doubled] (1.center) to (0.center);
		\draw [style=doubled] (0.center) to (3.center);
		\draw [style=doubled] (3.center) to (2.center);
		\draw [style=doubled] (2.center) to (1.center);
		\draw [style=doubled] (12.center) to (8.center);
		\draw [style=doubled] (6.center) to (10.center);
		\draw [style=doubled] (9.center) to (5.center);
		\draw [style=doubled] (11.center) to (7.center);
		\draw [style=doubled] (16.center) to (15.center);
		\draw [style=doubled] (15.center) to (14.center);
		\draw [style=doubled] (14.center) to (16.center);
		\draw [style=doubled] (20.center) to (18.center);
		\draw [style=doubled] (21.center) to (19.center);
		\draw [style=doubled] (23.center) to (22.center);
		\draw [style=doubled, bend left=15] (17.center) to (19.center);
		\draw [style=doubled] (31.center) to (30.center);
		\draw [style=doubled] (30.center) to (29.center);
		\draw [style=doubled] (29.center) to (31.center);
		\draw [style=doubled] (34.center) to (33.center);
		\draw [style=doubled] (37.center) to (36.center);
		\draw [style=doubled, bend right=15] (32.center) to (41);
		\draw [style=doubled] (47.center) to (45.center);
		\draw [style=doubled] (51.center) to (49.center);
		\draw [style=doubled] (56.center) to (54.center);
		\draw [style=doubled] (60.center) to (58.center);
		\draw [style=doubled] (62.center) to (61.center);
		\draw [style=doubled] (64.center) to (63.center);
		\draw [style=doubled, in=165, out=-90] (66.center) to (65.center);
		\draw [style=doubled, in=15, out=-90] (69.center) to (70.center);
		\draw [style=doubled, in=15, out=-90] (73.center) to (74.center);
		\draw [style=doubled, in=165, out=-90] (78.center) to (77.center);
		\draw [style=doubled] (84.center) to (83.center);
		\draw [style=doubled] (83.center) to (82.center);
		\draw [style=doubled] (82.center) to (84.center);
		\draw [style=doubled] (87.center) to (86.center);
		\draw [style=doubled] (90.center) to (89.center);
		\draw [style=doubled] (96.center) to (95.center);
		\draw [style=doubled] (98.center) to (97.center);
		\draw [style=doubled, in=165, out=-90] (100.center) to (99.center);
		\draw [style=doubled, in=15, out=-90] (103.center) to (104.center);
		\draw [style=doubled, in=165, out=-90] (110.center) to (109.center);
		\draw [style=doubled, in=15, out=-90] (113.center) to (114.center);
	\end{pgfonlayer}
\end{tikzpicture}}
\end{equation}
Example \ref{eq:MainSentence} can be displayed in this framework as well. 
The sentence $\sigma$ is a probability distribution over the two different realisations of the circuit, as in Equation \ref{eq:dogBrokeVaseDisCoCirc}. 
\begin{equation}
\label{eq:dogBrokeVaseDisCoCirc}
    \input{dogBrokeVaseDisCoCirc.tikz}
\end{equation}
With the above, the proposed theory of modelling probability distributions over different syntactic sentences is embedded in the DisCoCirc framework. 






\section{Implementation}
\label{sec:implementation}
We conduct a preliminary demonstration,
which yields the final contribution of this paper. 
The dataset consists of ten words (inspired by \citet{pronounResolution}): 
\[
    \texttt{pancakes, pasta, women, men, tasty, delicious, hungry, starving, are, is}
\]
Using this vocabulary, all possible meaningful combinations of words to sentences are composed, assigned the label \texttt{True}, and the non-meaningful combinations of words are composed, assigned the label \texttt{False}\footnote{The combinations resulting from the words \texttt{women}, \texttt{men} and \texttt{tasty} are labelled as \texttt{False}}. The sentences in this dataset are:
\begin{center}
\begin{tabular}{lllll}
    \texttt{pancakes are hungry} & \texttt{False} & \quad & \texttt{pancakes are starving} & \texttt{False} \\
    \texttt{pancakes are tasty} & \texttt{True} & \quad & \texttt{pancakes are delicious} & \texttt{True} \\
    \texttt{pasta is hungry} & \texttt{False} & \quad & \texttt{pasta is starving} & \texttt{False} \\
    \texttt{pasta is tasty} & \texttt{True} & \qquad \qquad & \texttt{pasta is delicious} & \texttt{True} \\
    \texttt{women are tasty} & \texttt{False} & \quad & \texttt{women are delicious} & \texttt{False}
    \\
    \texttt{women are starving} & \texttt{True} & \quad & \texttt{women are hungry} & \texttt{True} \\
    \texttt{men are tasty} & \texttt{False} & \quad & \texttt{men are delicious} & \texttt{False}
    \\
    \texttt{men are starving} & \texttt{True} & \quad & \texttt{men are hungry} & \texttt{True} 
\end{tabular}
\end{center}
In total, the dataset consists of \num{16} sentences, with eight sentences each labelled \texttt{False} and \texttt{True}. 

\citet{qnlpInPractice} used the \verb|Tket| compiler \citep{tket}, integrated into \verb|Lambeq| to simulate quantum hardware on a classical computer.
The \texttt{Tket} model closely resembles a quantum computer and uses \verb|pytket|\footnote{\url{https://pypi.org/project/pytket/} (accessed 12.08.2024)} to perform noisy, {architecture-aware}, \textit{shot-based} simulations of a quantum computer, which can be run on real quantum hardware. The term \textit{shot-based} refers to running the model numerous times to obtain an estimate of the probability distribution. 

\begin{figure}
    \centering
    \includegraphics[width=0.85\linewidth]{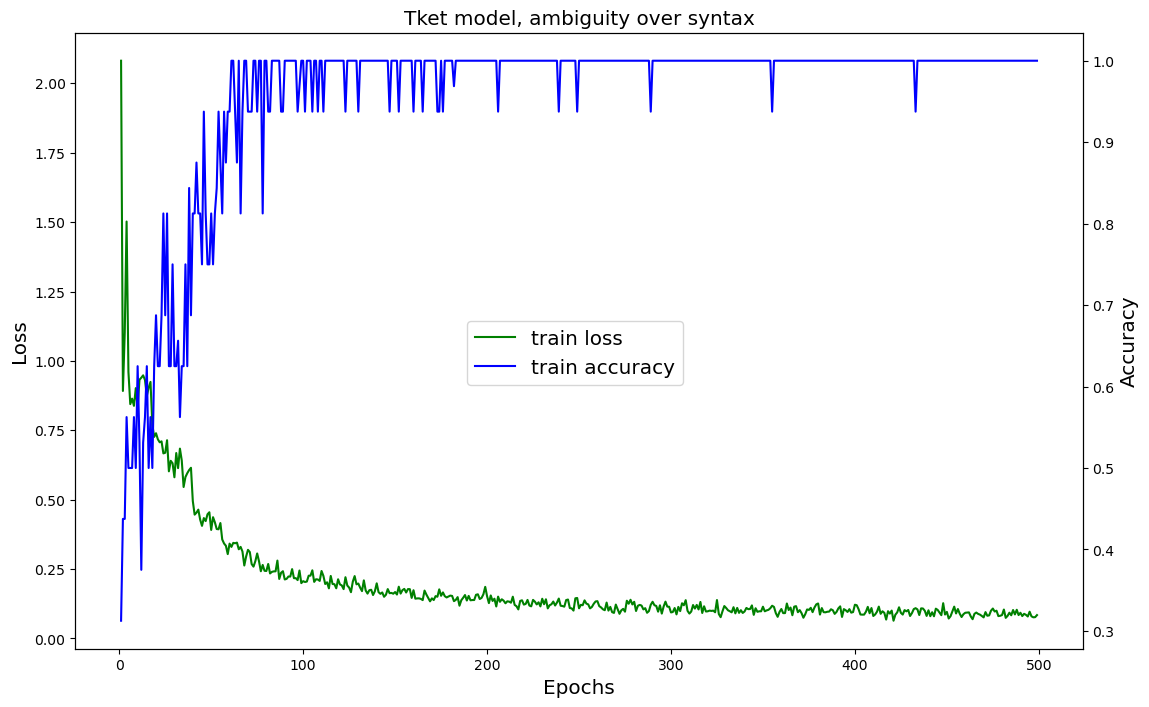}
    \caption{The train loss and train accuracy curves for the \texttt{Tket} model, trained on the dataset introduced in the main text, noun-meaning encoded on one qubit, final accuracy on the training set: \num{1.0}, $\kappa$ = \num{1.0}, $f$1-score = \num{1.0}}
    \label{fig:syntaxAmbLoss}
\end{figure}

\begin{figure}[h]
    \centering
    \resizebox{0.7\textwidth}{!}{\begin{tikzpicture}
	\begin{pgfonlayer}{nodelayer}
		\node [style=none] (104) at (-7.75, 4) {};
		\node [style=none] (105) at (-8.5, 4) {\large $\ket{0}$};
		\node [style=none] (106) at (-7.75, 4) {};
		\node [style=none] (107) at (-7.75, 2) {};
		\node [style=none] (108) at (-8.5, 2) {\large $\ket{0}$};
		\node [style=none] (109) at (-7.75, 2) {};
		\node [style=none] (110) at (-7.75, 0) {};
		\node [style=none] (111) at (-8.5, 0) {\large $\ket{0}$};
		\node [style=none] (113) at (-7.75, -2) {};
		\node [style=none] (114) at (-8.5, -2) {\large $\ket{0}$};
		\node [style=none] (115) at (-7.75, -4) {};
		\node [style=none] (116) at (-8.5, -4) {\large $\ket{0}$};
		\node [style=none] (118) at (-11, 0.25) {\texttt{noun}};
		\node [style=none] (119) at (-11, -0.25) {\texttt{one}};
		\node [style=none] (120) at (-11, -1.75) {\texttt{noun}};
		\node [style=none] (121) at (-11, -2.25) {\texttt{two}};
		\node [style=none] (122) at (-12.25, 3) {\texttt{adjective}};
		\node [style=none] (123) at (-9.75, 3) {\large  $\Biggl\{$};
		\node [style=none] (124) at (-7.25, -4) {};
		\node [style=none] (125) at (8, -4) {};
		\node [style=hadamard, minimum width=6mm, minimum height=6mm] (126) at (-4.75, -4) {\large $H$};
		\node [style=none] (127) at (-7.25, -2) {};
		\node [style=none] (128) at (8.25, -2) {};
		\node [style=hadamard, minimum width=6mm, minimum height=6mm] (129) at (-4.75, -2) {\large $U_{\texttt{noun two}}$};
		\node [style=none] (130) at (-7.25, 0) {};
		\node [style=none] (131) at (8.25, 0) {};
		\node [style=hadamard, minimum width=6mm, minimum height=6mm] (132) at (-4.75, 0) {\large $U_{\texttt{noun one}}$};
		\node [style=none] (133) at (-7.25, 2) {};
		\node [style=none] (134) at (-6.75, 2) {};
		\node [style=none] (136) at (-7.25, 4) {};
		\node [style=none] (137) at (-6.75, 4) {};
		\node [style=none] (138) at (-2.75, 2) {};
		\node [style=none] (139) at (8.25, 2) {};
		\node [style=none] (140) at (-2.75, 4) {};
		\node [style=none] (141) at (8.75, 4) {};
		\node [style=none] (142) at (-6.75, 4.75) {};
		\node [style=none] (143) at (-2.75, 4.75) {};
		\node [style=none] (144) at (-6.75, 1.25) {};
		\node [style=none] (145) at (-2.75, 1.25) {};
		\node [style=none] (146) at (-4.75, 3) {\large $U_{\texttt{adjective}}$};
		\node [style=none] (147) at (8.25, 2) {};
		\node [style=none] (148) at (8.25, 0) {};
		\node [style=none] (149) at (7.75, 2) {};
		\node [style=hadamard, minimum width=6mm, minimum height=6mm] (150) at (8.25, 2) {$\not \frown$};
		\node [style=hadamard, minimum width=6mm, minimum height=6mm] (151) at (8.25, 0) {$\not \frown$};
		\node [style=none] (152) at (8.25, -2) {};
		\node [style=hadamard, minimum width=6mm, minimum height=6mm] (153) at (8.25, -2) {$\not \frown$};
		\node [style=vertex] (155) at (-0.75, -4) {};
		\node [style=vertex] (156) at (-0.75, -2) {};
		\node [style=none] (157) at (-0.75, 2) {$\bigoplus$};
		\node [style=vertex] (158) at (3.25, -4) {};
		\node [style=vertex] (159) at (3.25, 0) {};
		\node [style=none] (160) at (3.25, 2) {$\bigoplus$};
		\node [style=hadamard, minimum width=7mm, minimum height=6mm] (161) at (1.25, -4) { $X$};
		\node [style=hadamard, minimum width=6mm, minimum height=6mm] (163) at (5.75, 0) {\large $H$};
		\node [style=hadamard, minimum width=6mm, minimum height=6mm] (164) at (5.75, -2) {\large $H$};
		\node [style=rightground] (165) at (8.25, -4) {};
	\end{pgfonlayer}
	\begin{pgfonlayer}{edgelayer}
		\draw (107.center) to (109.center);
		\draw (124.center) to (125.center);
		\draw (127.center) to (128.center);
		\draw (130.center) to (131.center);
		\draw (133.center) to (134.center);
		\draw (136.center) to (137.center);
		\draw (138.center) to (139.center);
		\draw (140.center) to (141.center);
		\draw (142.center) to (144.center);
		\draw (144.center) to (145.center);
		\draw (145.center) to (143.center);
		\draw (143.center) to (142.center);
		\draw (149.center) to (147.center);
		\draw (150) to (147.center);
		\draw (157.center) to (156);
		\draw (156) to (155);
		\draw (160.center) to (159);
		\draw (159) to (158);
	\end{pgfonlayer}
\end{tikzpicture}}
    \caption{The circuit used to create a probability distribution over the possible predictions of the model. The sentence meaning is encoded onto the upper qubit, and the lower qubit is discarded}
    \label{diag:implementationCircuit}
\end{figure}
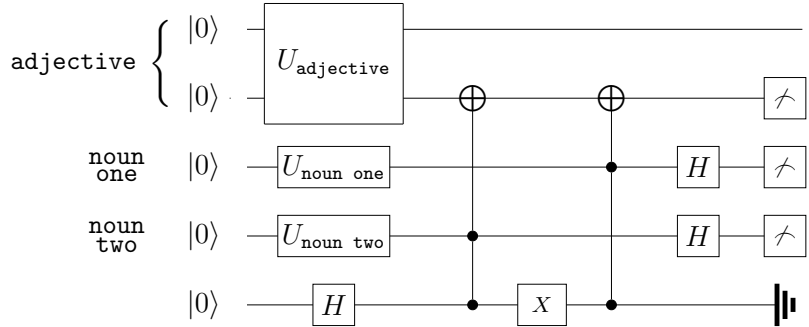

We train a \texttt{Tket} model\footnote{The code and dataset can be accessed at \url{https://github.com/jurekjurek/ManagingAmbiguity}} on the task of matching the resulting quantum circuits to labels (where we choose $\texttt{True} = (1, 0)^T$ and $\texttt{False} = (0, 1)^T$). One qubit to encode the noun meanings and use the \texttt{IQP}-ansatz with the same parameters as \citet{qnlpInPractice, pronounResolution}. 
In this preliminary study, only training and testing set are used, without a validation set. Given the size of the dataset (\num{16} sentences), introducing a validation set would substantially lower the model's ability to optimise the parameters. 
The resulting loss curve is shown in Figure \ref{fig:syntaxAmbLoss}. 
The accuracy, Cohen's kappa \citep{cohenKappa}, as well as the $f$1-score \citep{fMeasure} are \num{1.0}.
Now, a quantum circuit that creates a probability distribution over two different predictions is constructed (Figure \ref{diag:implementationCircuit}). 
This quantum circuit encodes probability distributions over predicting \texttt{False} or \texttt{True} with a respective probability of \num{0.5} (the magnitudes of the respective probabilities come from the \texttt{Hadamard} gate). 


We create all possible combinations of \texttt{noun} \texttt{one} and \texttt{noun} \texttt{two}, where they have to belong to different categories, e.g., \texttt{men} and \texttt{pancakes}, but not \texttt{pancakes} and \texttt{pasta}. 
We do this to obtain only probability distributions over {different} predictions of the model. 
As such, we obtain one sentence that is to be predicted as \texttt{False} and one sentence that is to be predicted as \texttt{True} and thus expect a probability distribution over the quantum states representing the label \texttt{True} and representing the label \texttt{False}. 
We then, for all of these combinations, capture the value of the {entropy} (Equation \ref{eq:vonNeumannEntropy}) of the resulting density matrix (encoded on the upper qubit in Figure \ref{diag:implementationCircuit}) and the values of the {fidelity} (Equation \ref{eq:fidelity}). The values of the fidelity are determined by comparison with the density matrices $\rho_{\texttt{True}}$ and $\rho_{\texttt{False}}$, where: 
\begin{equation}
\label{eq:trueFalseMatrix}
    \rho_{\texttt{True}} = \begin{pmatrix}
        1 & 0 \\ 0& 0
    \end{pmatrix} \qquad \qquad \rho_{\texttt{False}} = \begin{pmatrix}
        0& 0 \\ 0& 1
    \end{pmatrix} 
\end{equation}
For each combination of words, we obtain one value for the entropy and two values for the fidelity. 
Note that, given the explicit matrix representations in Equation \ref{eq:trueFalseMatrix}, the density matrix we observe would, in the optimal case be a probability distribution over these two {perfect} predictions, which is: 
\begin{equation*}
    \rho_{\texttt{optimal}} = \frac{1}{2}  \begin{pmatrix}
        1& 0 \\ 0& 1
    \end{pmatrix} 
\end{equation*}
The entropy for this optimal density matrix is one, and the fidelity, taken with respect to either of the pure matrices in Equation \ref{eq:trueFalseMatrix}, is \num{0.5}. 
However, since the model does not make perfect predictions, this result will not be achieved. 
The reason why the model does not make perfect predictions although the accuracy the model converged to is \num{1} is that the model's quality of predictions is based on the ultimate \textit{category} it predicts, not the probability distribution. So while the category might be correct, the probability distribution over the categories might not be perfectly separated, resulting in the model making wrong predictions in some of the cases, due to the statistical nature of the process. 

As a last step to obtaining the resulting values for the entropy and fidelity, their values for the different combinations of words are averaged.

We use \texttt{Qiskit} \citep{qiskit} to create and manipulate density matrices using the trained models. 

\subsection{Results}

We report an average entropy of \num{0.642}. The optimal entropy value is \num{1}, because only combinations of words that are of different kinds are considered, meaning that the resulting probability distribution is one over the model predicting \texttt{True} and the model predicting \texttt{False}. We would only be able to observe the optimal entropy value in the case of the two \textit{pure} states being perfectly separated. The pure density matrices overlap, as the model's predictions are not optimal. 
We report an average fidelity with the density matrix $\rho_{\texttt{True}}$ of \num{0.694}, and the fidelity with the density matrix $\rho_{\texttt{False}}$ is \num{0.306}. Note that the optimal fidelity value is \num{0.5} for both of these cases, as mentioned above.
We argue that this suboptimal balance is due to the {pure states} representing the meanings of, e.g., \texttt{Pancakes are hungry} or \texttt{Men are hungry}, not being perfectly predicted by the model. 
Ultimately, we do not see a perfect probability distribution over the states $\rho_{\texttt{True}}$ and $\rho_{\texttt{False}}$, but rather the probability distribution over the {predictions} of the model, which are imperfect. 

To quantify how imperfect the model's predictions are, the fidelity between the pure states representing the predictions of the model, and the density matrices $\rho_{\texttt{True}}$ and $\rho_{\texttt{False}}$ is considered. 
 
We record an average fidelity of \num{0.874} for the \texttt{True} sentences with the density matrix $\rho_{\texttt{True}}$, and an average fidelity of \num{0.719} for the \texttt{False} sentences with the density matrix $\rho_{\texttt{False}}$. 
As for the fidelity of the predicted \texttt{False} states with the density matrix $\rho_{\texttt{True}}$, we find a value of \num{0.281}, and the predicted \texttt{True} states with the density matrix $\rho_{\texttt{False}}$, we find a value of \num{0.126}. 

\subsection{Discussion}

The model's predictions, which yield \texttt{True}, have a bigger overlap with the correct density matrix than the predictions that yield \texttt{False}. 
This means that the model makes the predictions when it comes to \texttt{True} sentences more reliably. 
The model's prediction is thus a probability distribution over two decisions. 
Given that the model predicts \texttt{True} labels more correctly, it makes sense that 
the model is more likely to predict the label \texttt{True} which explains the higher fidelity with the matrix $\rho_{\texttt{True}}$ and the entropy lower than 1 for the whole circuit. 
Our findings suggest that the probability distribution over different wirings in DisCoCat diagrams can be modelled on a quantum computer and that one can reason with the density matrices that capture these probability distributions with the help of quantum circuits.



\section{Conclusion and Further Work}
\label{sec:conclusionPaper}
In this paper, we proposed a more natural way of reasoning with quantum linguistic models about certain cases of ambiguity. 
We present a quantum circuit architecture that creates a classical probability distribution over different operations being carried out, subsequently creating a density matrix. 

The arising quantum circuits can be interpreted to express sentence meaning, which can be expressed in terms of DisCoCat diagrams. The arising quantum circuits represent probability distributions over sentences that can contain different syntax. 

We give several example applications of the proposed theory, where we explicitly construct the quantum circuits as probability distributions over sentences. 
We relate the proposed theory to the DisCoCirc framework, where we philosophise about the introduction of \emph{dynamic verbs}: verbs whose meanings evolve in the text, and lastly, we conduct an experiment. 
In the experiment, we use a small dataset to create probability distributions over different sentence meanings, which are either \texttt{True} or \texttt{False}. 

Further work entails more involved experiments with larger datasets. 
Furthermore, one might tackle binary classification tasks such as \citet{qnlpInPractice, pronounResolution}, where one recovers the correct wiring of the words not in a syntax-based, but in a semantics-based approach. 

One might consider how exactly ambiguous words interact to give ambiguous meanings of sentences. Although a substantial amount of research still is to be done on this, consider the sentence: 
\begin{equation*}
    \texttt{The old man trains.}
\end{equation*}
This ambiguous sentence (the sentence can either mean \texttt{Old people take control of trains} or \texttt{The man, who is old, trains}) arises due to the individual words having ambiguous meanings. Modeling exactly how these individual ambiguities interact to give the ambiguous sentence meaning will be a focus of further research. 



\section*{Contributions}
JE conducted the research under the supervision of WG, LH and GW. GW proposed the project. JE wrote the text and the other authors contributed to editing and finalising it.

\section*{Acknowledgements}
This research received funding from the Flemish Government under the ``Onderzoeksprogramma Artifici\"ele Intelligentie (AI) Vlaanderen'' programme.

\bibliographystyle{apalike}
\bibliography{Central}
\end{document}